\pdfoutput=1

\documentclass[11pt]{article}

\usepackage{ACL2023}

\usepackage{times}
\usepackage{latexsym}

\usepackage{graphicx}
\usepackage{caption}
\usepackage{algorithm}
\usepackage{algorithmic}
\usepackage{multirow}
\usepackage{amsmath}
\usepackage{tabularx}
\usepackage{subcaption}
\usepackage{booktabs}

\usepackage[T1]{fontenc}

\usepackage[utf8]{inputenc}

\usepackage{microtype}

\usepackage{inconsolata}

\DeclareMathOperator*{\argmax}{argmax}

\title{Easy Guided Decoding in Providing Suggestions \\
for Interactive Machine Translation}

\author{Ke Wang\thanks{~~indicates equal contribution.}, Xin Ge$^{*}$, Jiayi Wang, Yu Zhao, Yuqi Zhang\thanks{~~indicates the corresponding author.} \\
Alibaba Group Inc. \\ \ \ \ 
\texttt{\{moyu.wk,shiyi.gx,joanne.wjy,kongyu, chenwei.zyq\}@alibaba-inc.com} \\ 
}

\begin{document}
\maketitle
\begin{abstract}
Machine translation technology has made great progress in recent years, but it cannot guarantee error-free results. Human translators perform post-editing on machine translations to correct errors in the scene of computer-aided translation. In favor of expediting the post-editing process, many works have investigated machine translation in interactive modes, in which machines can automatically refine the rest of translations constrained by human’s edits. Translation Suggestion (TS), as an interactive mode to assist human translators, requires machines to generate alternatives for specific incorrect words or phrases selected by human translators.  In this paper, we utilize the parameterized objective function of neural machine translation (NMT) and propose a novel constrained decoding algorithm, namely Prefix-Suffix Guided Decoding (PSGD), to deal with the TS problem without additional training. 
Compared to the state-of-the-art lexically constrained decoding method, PSGD improves translation quality by an average of $10.87$ BLEU and $8.62$ BLEU on the \textit{WeTS}\footnote{\url{https://github.com/ZhenYangIACAS/WeTS}} and the WMT 2022 Translation Suggestion datasets\footnote{\url{https://www.statmt.org/wmt22/translation-suggestion-task.html}}, respectively, and reduces decoding time overhead by an average of $63.4\%$ tested on the WMT translation datasets. Furthermore, on both of the TS benchmark datasets, it is superior to other supervised learning systems trained with TS annotated data.
\end{abstract}

\section{Introduction and Related Work}

The emergence of machine translation technology \citep{lopez2008statistical,koehn2009statistical} assists human translation to improve translation efficiency \citep{green2014predictive,green2015natural,herbig2020mmpe}. Even though there is a quality gap between the outputs of machine translation (MT) systems and manual translations by professional translators, MT can still practically reduce time in comparison with translating from scratch \citep{laubli2013assessing}.  Later, the advances at the sequence-to-sequence model \citep{sutskever2014sequence,bahdanau2014neural,vaswani2017attention} further made a breakthrough in translation technology, inspiring the industry to transform human translation into computer-aided translation (CAT) \citep{knowles-koehn-2016-neural,santy2019inmt} to a great extent. CAT usually relies on an MT engine and a platform with a user-friendly interface \citep{bowker2002computer,lengyelmemoq,bowker2010computer,bowker2014computer,pal2016catalog,chatterjee2019automatic}, with which humans perform post-editing (PE) on machine translations to achieve final results with quality standards. 

\begin{figure*}
\centering
\includegraphics[width=1.0\textwidth]{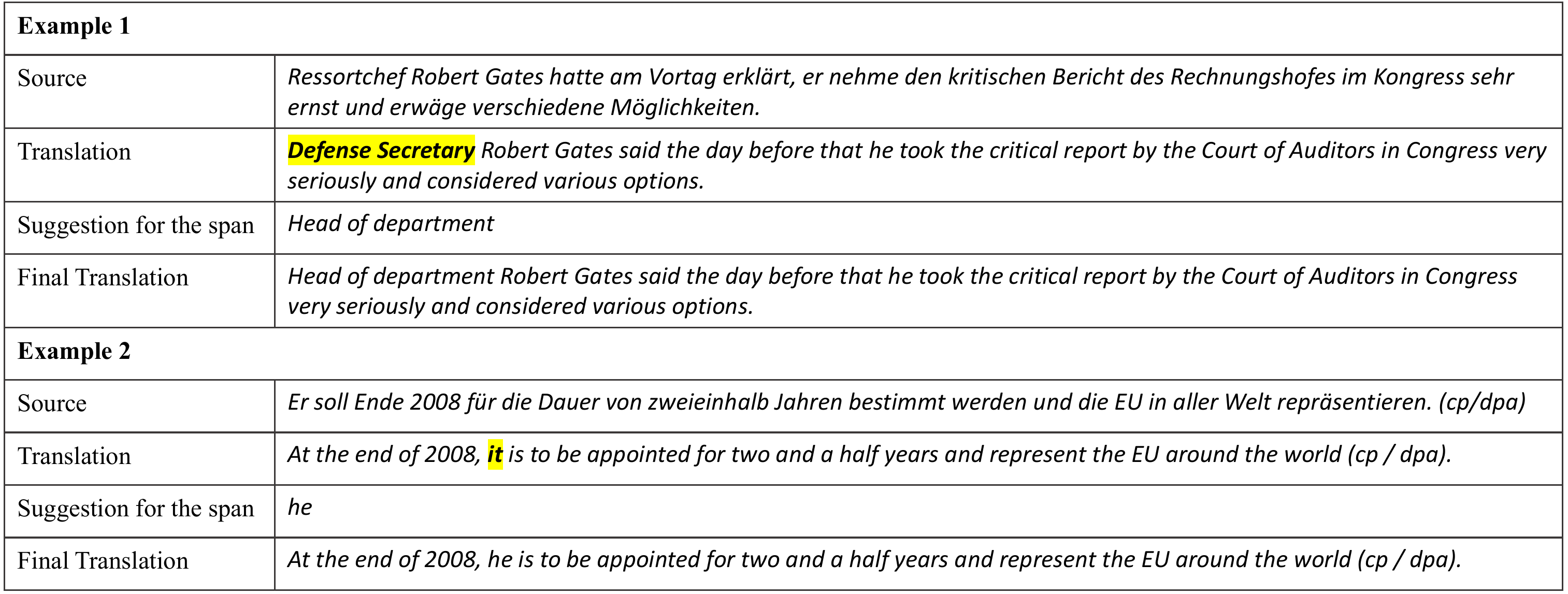} 
\caption{Examples of Translation Suggestions in computer-aided translation. Highlighted are spans with incorrect words selected by humans. The TS system automatically generates alternatives for the spans to obtain final translation results. }
\label{fig:span}
\end{figure*}

In the past, the post-editing process was typically static and machines would no longer respond to humans'  modifications once humans started post-editing. Recent works \citep{domingo-etal-2016-interactive,gonzalez-rubio-etal-2016-beyond,peris2017interactive} investigate interactive protocols and algorithms so that humans and machines can collaborate and machines automatically refine the translations according to the human's edits. One promising mode is Translation Suggestion (TS) proposed by \citet{yang2021wets} as a pioneer, which requires the machine to provide alternatives for specific spans of incorrect words or phrases selected by humans, namely making suggestions given prefix and suffix constraints. In practical applications as shown in Figure \ref{fig:span}, it usually happens when human translators would like to edit part of the MT output. It can be easily implemented with a user interface if machines correctly provide suggestions for the selected incorrect spans. \citet{yang2021wets} has proven the significance of TS in post-editing in terms of resolving two pitfalls of earlier works. The importance has also been recognized by the Conference of Machine Translation (WMT), and they released the Naive Translation Suggestion shared task \cite{yang-EtAl:2022:WMT1} in WMT 2022\footnote{\url{https://statmt.org/wmt22/translation-suggestion-task.html}}. 

One of the solutions to TS can be training a supervised model with TS annotated data. \citet{yang2021wets} trained such an end-to-end Transformer-like model as a benchmark system. \citet{ge2022tsmind} applied model fine-tuning with TS data augmentation on pre-trained NMT models. However, supervised learning, which relies on a large amount of labeled data, is too heavy to be easily adjusted to other domains. In addition, due to the complicated post-editing process, it is expensive to obtain such limited annotated data.

Our idea is to investigate inference algorithms given prefix and suffix constraints. 
We tested the state-of-the-art lexically constrained decoding algorithm \citep{post-vilar-2018-fast,hu-etal-2019-improved} on the \textit{WeTS} dataset, and found that omissions frequently occur in suggestion generations. There are two reasons behind: (1) the division of beams in the dynamic beam allocation narrows the search space so that there would not be enough candidates that match constraints to be picked. This is more likely to happen when constraints are much longer than the average length of the span selected, such as in TS applications; (2) the beam search stops when the probability of \textit{eos}, the special token for the end of a sentence, appears largest in the softmax distribution, but the probability of the entire sentence generation has not been considered. In terms of efficiency, this decoding algorithm contains unnecessary calculations since it always generates suggestions step by step from the beginning of the translation to the end of the sentence, including prefix and suffix constraints.

In this paper, we propose a neat prefix-suffix guided decoding (PSGD) algorithm for the TS task. There are three main contributions: (1) PSGD emphasizes the probability of the entire generation including prefix and suffix constraints, but only decodes for the incorrect span rather than the whole translation sentence, which improves both suggestion quality and time efficiency. (2) PSGD theoretically avoids dividing beams as in \citet{hu-etal-2019-improved} so that the original beam search space can be used to improve the quality of generated translation suggestions. (3) PSGD does not require any additional training/fine-tuning on the original NMT model, which means it can be applied to any auto-regressive machine translation system with flexibility.


Our experimental observations show that PSGD significantly outperforms the state-of-the-art lexically constrained decoding method \citep{post-vilar-2018-fast,hu-etal-2019-improved} by an average increase of $10.87$ BLEU and $8.62$ BLEU on the benchmark datasets \textit{WeTS} and WMT 2022 Naive Translation Suggestion datasets (\textit{WMT22-TS}), respectively. Experimental results also demonstrate PSGD's superiority in overall time efficiency by a $63.4\%$ time reduction. In addition, on both the \textit{WeTS} and \textit{WMT22-TS} datasets, PSGD is superior over other supervised learning systems trained with TS annotated data.

\section{Preliminary}

\begin{table}[t]
  \begin{tabularx}{1.0\linewidth}{c|X}
    \toprule
    Symbol & Definition\\
    \midrule
    $\mathbf{x}$ & Sentence in source language \\  \hline
    $\mathbf{y}$ & Translated sentence in target language \\  \hline
    $\Theta$ & Model parameters \\ \hline
    $\mathbf{p}$ & A given prefix of  $\mathbf{y}$ \\ \hline
    $\mathbf{s}$ & A given suffix of  $\mathbf{y}$ \\  \hline
    $\mathbf{r}$ & The remained part of $\mathbf{y}$ after $\mathbf{p}$ and $\mathbf{s}$ are removed \\  \hline
    $t_p$ & The number of tokens in $\mathbf{p}$ \\ \hline
    $t_s$ & The number of tokens in $\mathbf{s}$ \\ \hline
    $t_r$ & The number of tokens in $\mathbf{r}$ \\ \hline
    {\it bos} & The special begin of sentence token used in machine translation \\ \hline
    {\it eos} & The special end of sentence token used in machine translation \\ \hline
    $y_i$ & The $i$-th token of $\mathbf{y}$, similar for $x_i,p_i,s_i$,$r_i$, etc. \\ \hline
    $\hat{\mathbf{y}}$ & The estimation/prediction of $\mathbf{y}$, similar for $\hat{\Theta}$, $\hat{\mathbf{r}}$, $\hat{r}_j$, etc. \\ \hline
    $\mathbf{v}_{<t}$ & The first $t$ tokens of sequence $\mathbf{v}$ \\ \hline
    $\langle\mathbf{v}_1, ..., \mathbf{v}_n\rangle$  & Concatenation of sequences $\mathbf{v}_1, ..., \mathbf{v}_n$ \\ \hline
    $P_{\text{MT}}$ & The probabilistic model of machine translation \\ \hline
    $P_r^{(n)}$ & The probability of the entire sequence after the $n$-th decoding step for the remain part $r$ \\ \hline
    $pt$ & Patience for decoding early stopping \\ \hline
    $f(y_t)$ & The softmax probability distribution at the $t$-th decoding step \\
  \bottomrule
\end{tabularx}
  \caption{Notations}
  \label{tab:notation}
\end{table}

Before introducing the PSGD, we first elaborate on the problem in the scene of TS with the mathematical notations in Table \ref{tab:notation}. 
The sequence-to-sequence machine translation model is generally a conditional auto-regressive language model that follows the factorized probability distribution.
\begin{equation}
\label{eq:loss}
P_{\text{MT}}(\mathbf{y}|\mathbf{x}; \Theta) = \prod_i P_{\text{MT}}(y_i|\mathbf{y}_{<i}, \mathbf{x}; \Theta)
\end{equation}
Given training pairs of $(\mathbf{x},\mathbf{y})$, maximizing the probability in Formula \ref{eq:loss} will return an estimate of the model parameter, $\hat{\Theta}$. 

During inference, we are supposed to maximize the probability of the generated sequence:
\begin{equation}
\label{eq:trainObj}
    \hat{\mathbf{y}} = \argmax_\mathbf{y}P_{\text{MT}}(\mathbf{y}|\mathbf{x}; \hat{\Theta})
\end{equation}

However, the solution space $\{\mathbf{y}\}$ is infinite and impossible to search. Therefore, an auto-regressive decoding process is applied as an approximation. The decoder of the model decodes the translation step by step greedily\footnote{We do not involve beam-search here to simplify the equations, but in practice, all our experiments are conducted with beam-search.} as following until \textit{eos} is met:
\begin{equation}
\label{eq:oldDecodingStep}
\hat{y_i} = \argmax_y P_{\text{MT}}(y|\mathbf{\hat{y}}_{<i}, \mathbf{x}; \hat{\Theta})
\end{equation}

In the scene of TS, a span $\mathbf{r}$ is masked for suggestion generation, and thus we need to guarantee that $\mathbf{y}$ starts with the given prefix $\mathbf{p}$ and ends with the given suffix $\mathbf{s}$. Then, the problem becomes:
\begin{equation}
\label{eq:overallProb}
    \hat{\mathbf{r}} = \argmax_\mathbf{r}P_{\text{MT}}(\langle\mathbf{p},\mathbf{r},\mathbf{s}\rangle|\mathbf{x}; \hat{\Theta})
\end{equation}

\begin{figure*}[t]
\centering
\includegraphics[width=0.9\linewidth]{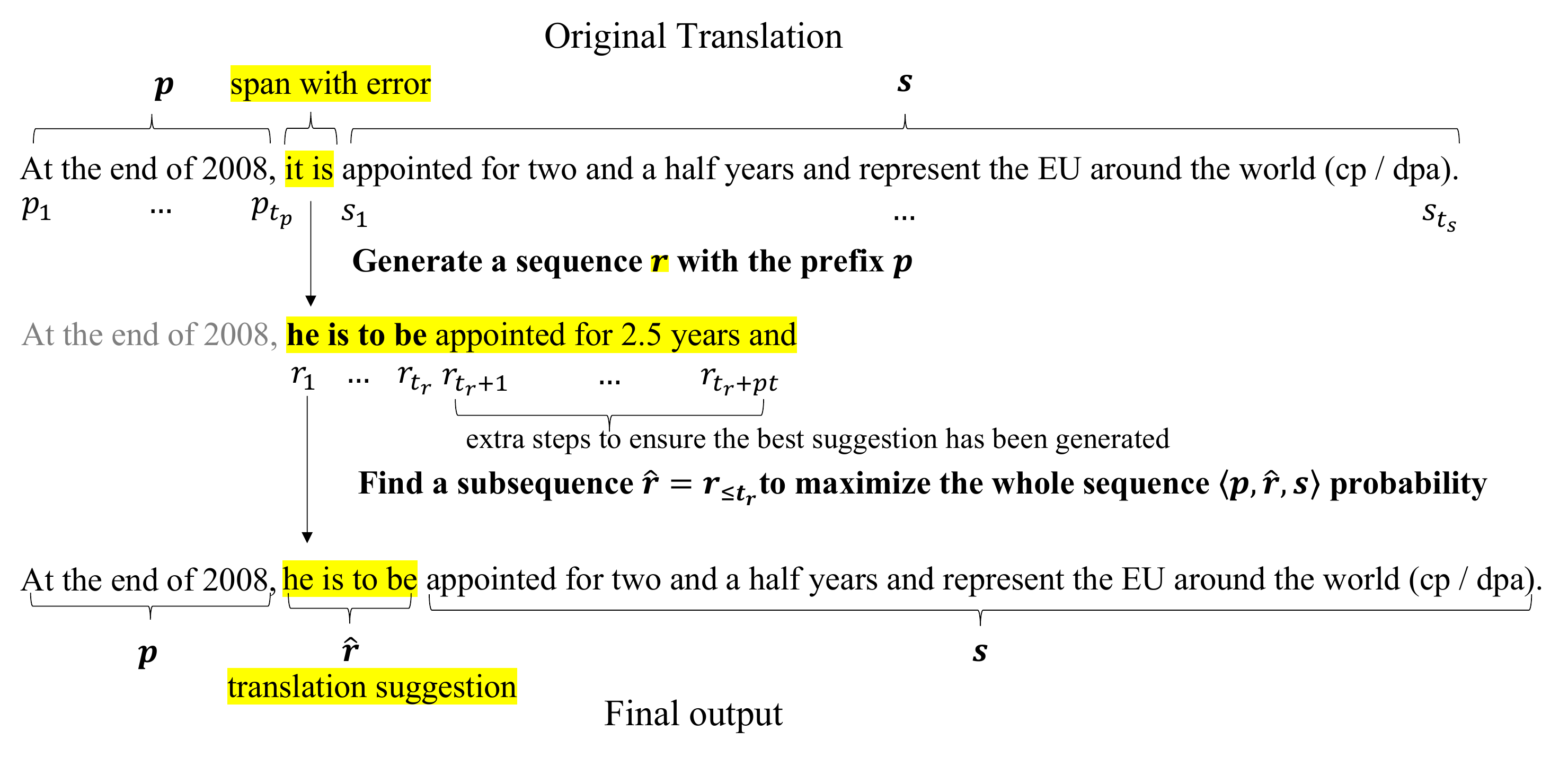} 
\caption{An example of applying the PSGD algorithm to generate translation suggestions}.
\label{fig:process}
\end{figure*}

\section{Methodology}
\label{method}
As mentioned previously, earlier works of constrained decoding require step-by-step generations for the whole sentence even with prefix and suffix constraints. It causes efficiency issues especially when the constraints are long. What's worse, the division of beams might bring about low-quality suggestions. Accordingly, PSGD generates translation suggestions given prefix and suffix constraints through maximizing the probability of the whole generated sequence. Meanwhile, PSGD performs a step-by-step beam search generation only for the span without dividing beams. A brief example of the PSGD algorithm for TS can be found in Figure \ref{fig:process}. With this picture in mind, we will give details of the PSGD algorithm in this Section. The pseudo codes of this algorithm are presented in Algorithm \ref{alg:decoding}. And the fairseq based implementation is available on github\footnote{\url{https://github.com/wangke1996/translation-suggestion-psgd}}. 

\begin{algorithm}[tb]
\caption{Algorithm of PSGD}
\label{alg:decoding}
\textbf{Input}: $\mathbf{x}$, $\mathbf{p}$, $\mathbf{s}$ \\
\textbf{Parameter}: $pt$ \\
\textbf{Output}: $\hat{\mathbf{r}}$
\begin{algorithmic}[1] 
\STATE Let $\hat{\mathbf{r}}=[]$, $n=0$, $maxP_r=-\infty$, $bestStep=0$.
\WHILE{$n-bestStep < pt$}
\STATE $\hat{\mathbf{y}} = \langle\mathbf{p},\hat{\mathbf{r}},\mathbf{s}\rangle$
\STATE $\{f(p_i)\},\{f(\hat{r}_j)\},\{f(s_k)\}=$ Output probability distribution of every tokens in $\hat{\mathbf{y}} = \langle\mathbf{p},\hat{\mathbf{r}},\mathbf{s}\rangle$
\STATE calculate $P_r^{(n)}$ with Formula \ref{eq:calculateWholePorb}
\IF {$P_r^{(n)} > maxP_r$}
\STATE $maxP_r = P_r^{(n)}$
\STATE $bestStep = n$
\ENDIF
\STATE calculate $\hat{r}_{n+1}$ with Formula \ref{eq:calculateNextCand}
\STATE append $\hat{r}_{n+1}$ to $\hat{\mathbf{r}}$
\STATE $n=n+1$
\ENDWHILE
\STATE \textbf{return} the first $bestStep$ elements of $\hat{\mathbf{r}}$
\end{algorithmic}
\end{algorithm}

\subsection{Decoding Process for TS}
Similar to the normal translation process, we apply the auto-regressive decoding process to overcome the issue of infinite solution space $\{\mathbf{r}\}$. We can factorize the entire probability in Formula \ref{eq:overallProb} into three parts:

\begin{align}
\label{eq:threeProd}
\begin{split}
& P_{\text{MT}}(\langle\mathbf{p},\mathbf{r},\mathbf{s}\rangle|\mathbf{x}; \Theta) \\ = & \prod_i P_{\text{MT}}(p_i|\mathbf{p}_{<i},\mathbf{x};\Theta)  \\ 
\cdot & \prod_j P_{\text{MT}}(r_j|\langle\mathbf{p}, \mathbf{r}_{<j}\rangle, \mathbf{x}; \Theta) \\
\cdot & \prod_k P_{\text{MT}}(s_k|\langle\mathbf{p}, \mathbf{r}, \mathbf{s}_{<k}\rangle, \mathbf{x}; \Theta)
\end{split}
\end{align}

Notice that in Formula \ref{eq:threeProd}, as all $p_i$ and $s_k$ are given, only the second product requires a step-by-step auto-regressive generation. 
The first product, referring to the probability of the prefix sequence $\mathbf{p}$, can be calculated at the first step when $r_1$ is decoded. And when all of the decoding steps of $\mathbf{r}$ are completed, we can easily obtain the third product, which refers to the probability of the suffix sequence $\mathbf{s}$. Hence, the decoding process at each step is actually:
\begin{equation}
\label{eq:constrainStep}
    \hat{r}_j = \argmax_r P_{\text{MT}}(r|\langle\mathbf{p},\mathbf{\hat{r}}_{<j}\rangle, \mathbf{x}; \hat{\Theta})
\end{equation}
Theoretically, we only need to go forward through the model for $t_r$ times, which is smaller than that of a normal constraint decoding \cite{hu-etal-2019-improved}, i.e. $t_p+t_r+t_s$. In particular, for the TS task, human translators might select a short span with incorrect words, and then $t_p+t_r+t_s$ could be much larger than $t_r$.

\subsection{Condition for Stopping Decoding}
The decoding process aforementioned is straightforward. Now, it comes to a problem: when should the inference stop? In \citet{post-vilar-2018-fast} and \citet{hu-etal-2019-improved}, the model generates the full sequence and the generation stops when the {\it eos} is met. However, in PSGD, only the span $\mathbf{r}$ will be generated and $\mathbf{r}$ does not end with {\it eos}. We cannot utilize the generation of {\it eos} as a signal to stop decoding.

The key to solving this problem can be aligned with the objective function of NMT training as described in Formula \ref{eq:loss}. We assume that we have finished decoding in Formula \ref{eq:constrainStep} for $N$ steps, where $N$ is large enough to obtain a hypothesis sequence ${\mathbf{r}}^*$. The solution space $\{\mathbf{r}\}$ becomes $\{\mathbf{r}^*_{<n}\}_{n=1} ^{N}$. The most optimal sequence $\hat{\mathbf{r}} = \mathbf{r}^*_{<\hat{n}}$ is obtained by calculating $\hat{n}$ as:

\begin{equation}
    \hat{n} = \argmax_n P_{MT}(\langle\mathbf{p},\mathbf{r}^*_{<n},\mathbf{s}\rangle|\mathbf{x};\hat{\Theta})
\end{equation}

The normal decoding process stops when {\it eos} is reached, and it does not necessarily mean that the probability of the entire sequence generation is maximized. In other words, there might be a better hypothesis if the model generates fewer or more steps in the normal decoding process. Our condition for stop decoding in PSGD approaches the training objective function in Formula \ref{eq:loss} as much as possible. This is the reason why PSGD achieves better performance in suggestions.

\subsubsection{Early Stopping}
Although the above workaround provides a proper condition for stopping decoding, it might cause efficiency issues since the number of decoding steps would become $N$ rather than $t_r$. Obviously, $N$ should be equal to or greater than $t_r$ to achieve the best performance. However, we have no idea how large $t_r$ is. We have to set $N$ to a large enough value to cover as many cases as possible, which is unacceptable in practice.

Instead of setting a large $N$, we try to apply an early stopping mechanism to save unnecessary decoding steps. In Formula \ref{eq:constrainStep}, after every decoding step, PSGD calculates and records the averaged probability of entire sequence generation with both prefix and suffix constraints concatenated, denoted as $P_{r}^{(n)}$ as follows:
\begin{equation}
    P_{r}^{(n)} = \frac{P_{MT}(\langle\mathbf{p},\hat{\mathbf{r}}_{<n},\mathbf{s}\rangle|\mathbf{x},\hat{\Theta})}{t_p+n+t_s}
\end{equation}
PSGD stops decoding when $P_{r}^{(n)}$ does not increase any more within $pt$ consecutive steps, where $pt$ represents patience for early stopping.

\begin{figure}[t]
\centering
\includegraphics[width=0.9\columnwidth]{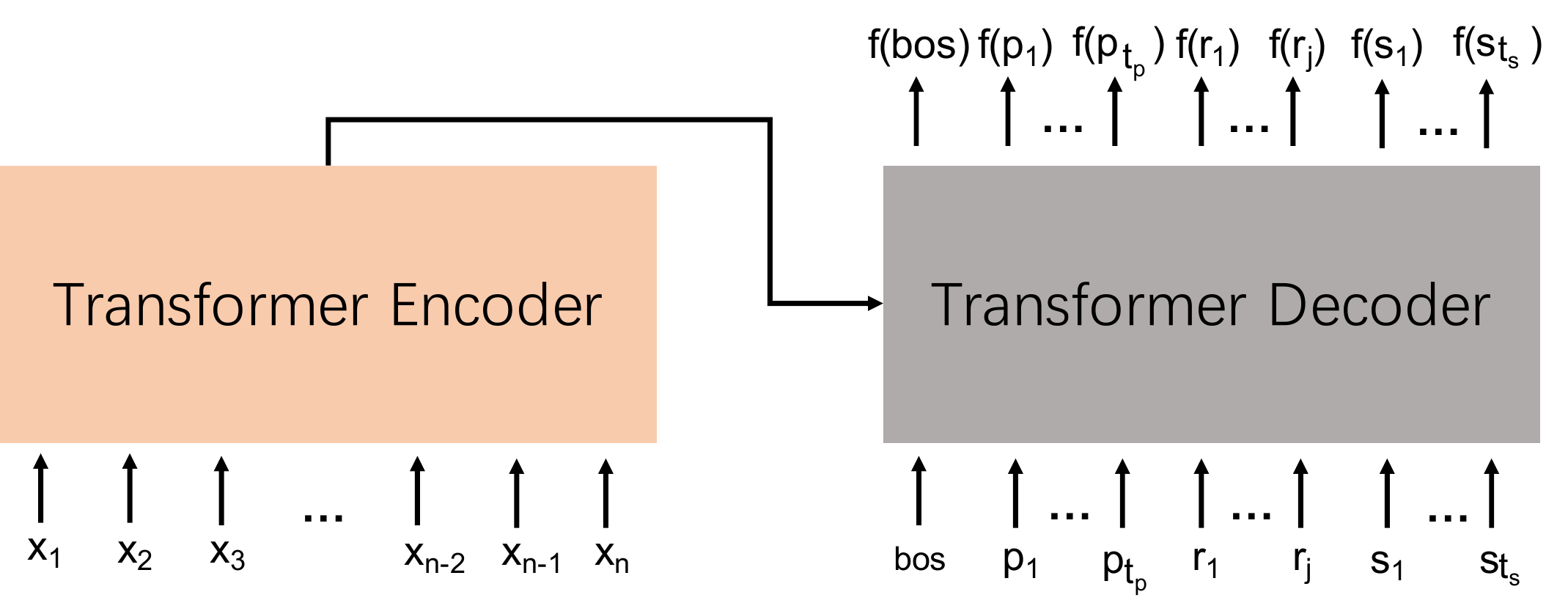} 
\caption{Forward path for calculating $P_{r}^{(n)}$}.
\label{fig:forward}
\end{figure}

\subsubsection{Parallel Acceleration}
Because the early stopping mechanism requires extra calculations for $P_{r}^{(n)}$ at each step, it seems that we would need to go forward through the model for $2*(t_r+pt)$ times. In fact, by utilizing the future mask in Transformer's decoder, we only run $t_r+pt$ times since $P_{r}^{(n)}$ and $\hat{r}_{n+1}$ can be obtained at the same time. Details are explained as follows. 

Figure \ref{fig:forward} shows the forward path for calculating $P_{r}^{(n)}$, where $f(\hat{y}_t)$ is the softmax probability distribution at the $t$-th decoding step. $f(y|y_t)$ indicates the probability of token $y$ at $(t+1)$-th step, which is short for $P_{MT}(y|\hat{\mathbf{y}}_{\leq t},\mathbf{x};\hat{\Theta})$. 
We have:
\begin{align}
\label{eq:calculateWholePorb}
\begin{split}
    P_r^{(n)} = & \frac{1}{t_p+n+t_s} \cdot f(p_1|{\it bos}) \cdot f(p_2|p_1) \cdot \dots \\
    \cdot & f(p_{t_p}|p_{t_p-1}) \cdot  f(\hat{r}_1|p_{t_p}) \cdot f(\hat{r}_2|\hat{r}_1) \cdot \dots \\
    \cdot & f(\hat{r}_n|\hat{r}_{n-1}) \cdot f(s_1|\hat{r}_n) \cdot f(s_2|s_1) \cdot \dots  \\
    \cdot & f(s_{t_s}|s_{t_s-1}) \cdot f({\it eos}|s_{t_s})
\end{split}
\end{align}
and
\begin{align}
\label{eq:calculateNextCand}
\begin{split}
    \hat{r}_{n+1} = & \argmax_r P_{\text{MT}}(r|\langle\mathbf{p},\mathbf{\hat{r}}_{\leq n}\rangle, \mathbf{x}) \\
    = & \argmax_r  f(p_1|{\it bos}) \cdot f(p_2|p_1) \cdot \dots \\
    \cdot & f(p_{t_p}|p_{t_p-1}) \cdot  f(\hat{r}_1|p_{t_p}) \cdot f(\hat{r}_2|\hat{r}_1) \cdot \\
      & \dots \cdot f(\hat{r}_n|\hat{r}_{n-1}) \cdot f(r|\hat{r}_n) \\
    = & \argmax_r  f(r|\hat{r}_n)
\end{split}
\end{align}
At the $n$-th step, we can simultaneously get the softmax distributions $\{f(p_i)\}$, $\{f(\hat{r}_j)\}$, and $\{f(s_k)\}$. Then, according to Formula \ref{eq:calculateWholePorb} and \ref{eq:calculateNextCand}, $P_r^{(n)}$ (the probability of the entire generation for early stopping) and $\hat{r}_{n+1}$ (the next token generation) can be obtained together. Therefore, the total step of the decoding process in PSGD is $t_r+pt$.

\subsection{Summary of Methodology}
In this section, we have provided full details of PSGD, and the pseudo code is presented in Algorithm \ref{alg:decoding}. By utilizing the parameterized objective function, PSGD maximizes the probability of the entire generation for better suggestion quality, but only generates the suggestion for the incorrect span instead of the whole sequence. Besides, it designs a proper early-stopping mechanism for efficiency improvement. Only $t_r+pt$ decoding steps are required, which is smaller than $t_p+t_r+t_s$ in \citet{hu-etal-2019-improved}, since $t_r$ is much smaller than $t_p+t_r+t_s$ on average in the scene of TS, and the value of the hyper-parameter $pt$ is proven to be quite small in the subsequent experimental observations.

\section{Experiments}

\subsection{Generation Performance on Translation Suggestion}

  



\begin{table*}[tbp]
\centering

\begin{tabular}{ccccccc}
\toprule
Method &\multicolumn{1}{c}{Note} & De-En          & En-De & Zh-En     & En-Zh     & Avg \\ \midrule
SA-Transformer \cite{yang2021wets} & Supervised  & 31.20         & 29.48                 &    25.51       &     36.28      &  30.62  \\ \midrule
VDBA \cite{hu-etal-2019-improved}  & \multirow{2}{*}{Unsupervised}                       & 23.71          & 31.29                      &     12.11      &      12.96     &   20.02  \\
PSGD (Ours) & & \textbf{34.74} & \textbf{38.92}             & \textbf{20.25} &\textbf{ 29.64} &  \textbf{30.89}   \\ \bottomrule
\end{tabular} 
\caption{BLEU results on the \textit{WeTS} test sets. PSGD outperforms VDBA by $+10.87$ BLEU on average, and it also outperforms the TS benchmark systems, SA-Transformer, trained with annotated data in German-English and English-German.}
\label{tab:ts_result}
\end{table*}

\begin{table*}[]
\centering
\begin{tabular}{ccccccc}
\toprule
Method                 & Note                          & De-En & En-De & En-Zh & Zh-En & Avg \\ \midrule
TSMind \cite{ge2022tsmind} & Supervised          & 33.23                      & 37.14                      & 21.20                      & 16.44                      & 27.00                    \\ \midrule
VDBA \cite{hu-etal-2019-improved}                 & \multirow{2}{*}{Unsupervised} & 23.16                      & 29.86                      & 13.37                      & 8.97                       & 18.84                    \\ 
PSGD (Ours)                   &                               & \textbf{33.87 }                     & \textbf{35.99}                      & \textbf{24.64}                      & \textbf{15.32}                      & \textbf{27.46}                    \\ \bottomrule
\end{tabular}
\caption{BLEU results on the \textit{WMT22-TS} test sets. PSGD outperforms VDBA by $+8.62$ BLEU on average, and it also slightly outperforms the winning system, TSMind, trained without pseudo corpus on average of 4 language pairs.}
\label{tab:wmt_result}
\end{table*}
Translation suggestion is a significant CAT task where human translators select spans with incorrect words or phrases in the MT results. It requires models to automatically provide alternatives for the spans for efficiency improvement in translation. \textit{WeTS} \citep{yang2021wets} and \textit{WMT22-TS} \citep{yang-EtAl:2022:WMT1} are the benchmark datasets for the TS task. Each sample consists of a source sentence, a masked MT sentence, and a golden reference corresponding to the masked span. The masked spans with potential translation errors are selected and annotated by humans. The annotations mainly focus on three types of translation errors: under (over)-translations, grammatical or syntactic errors, and semantic errors.

We compare the PSGD with three state-of-the-art systems on \textit{WeTS} and \textit{WMT22-TS}: SA-Transformer \cite{yang2021wets}, TSMind \cite{ge2022tsmind} (\textit{without TS data augmentation}) and VDBA \cite{hu-etal-2019-improved}. SA-Transformer is an end-to-end transformer-like model, which initializes the encoder with weights from the pre-trained XLM-Roberta \cite{xlmr} and fine-tunes it with TS labeled data. TSMind utilizes the pre-trained NMT model and fine-tunes it with TS triplets as well, and it wins in three out of four language pairs of the Naive Translation Suggestion task in WMT 2022 \citet{yang-EtAl:2022:WMT1}. VDBA is basically an improved version of the Dynamic Beam Allocation algorithm proposed in \citet{post-vilar-2018-fast}, that is remarkably capable in lexically constrained decoding. 

We should notice that both of SA-Transformer and TSMind perform supervised learning with human-annotated data for translation suggestion, while VDBA and our PSGD generate suggestions solely based on the original NMT model (released by \citet{yang2021wets}) without any additional training/fine-tuning on any TS-labeled data. Comparisons between SA-Transformer and TSMind have been reported in \citet{yang2021wets} and \citet{ge2022tsmind}, respectively. VDBA and PSGD are both implemented with the Fairseq toolkit \citep{ott2019fairseq} to generate suggestions. In evaluation, we calculate SacreBLEU metric \citep{post2018call} between the golden references for the masked spans and the alternatives generated by each system to measure their performances. 



\begin{figure}[t]
\centering
\includegraphics[width=0.9\columnwidth]{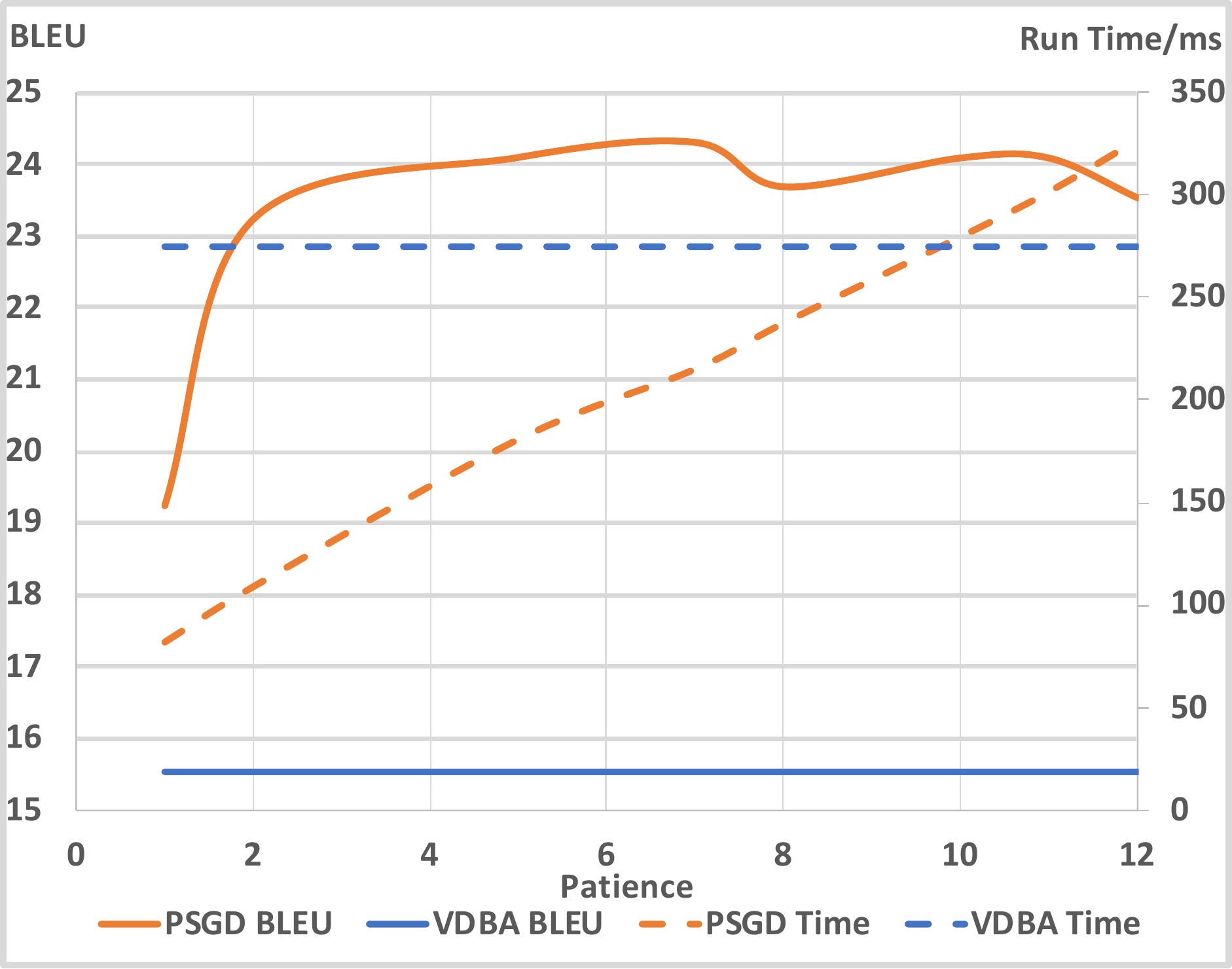} 
\caption{BLEU and decoding time cost with different early stopping patience ${pt}$ on the \textit{WeTS} German-English development set. }
\label{fig:early_stop_fig}
\end{figure}
\begin{figure}[ht]
    \centering
    \begin{subfigure}[t]{0.49\linewidth}
        \includegraphics[width=\linewidth]{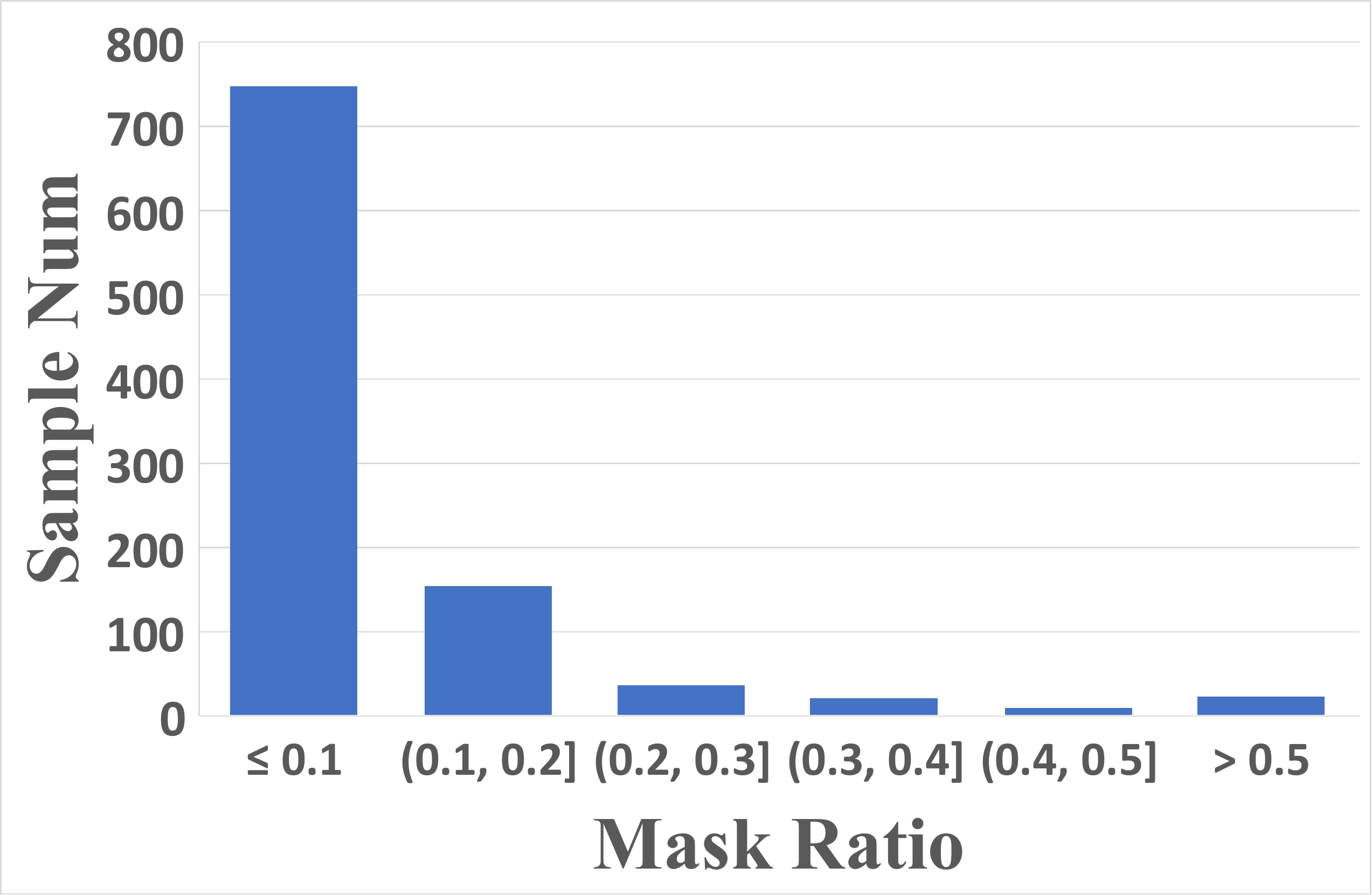}
        \caption{German to English}
        \label{fig:mask_percent:de2en} 
    \end{subfigure}
    \begin{subfigure}[t]{0.49\linewidth}
        \includegraphics[width=\linewidth]{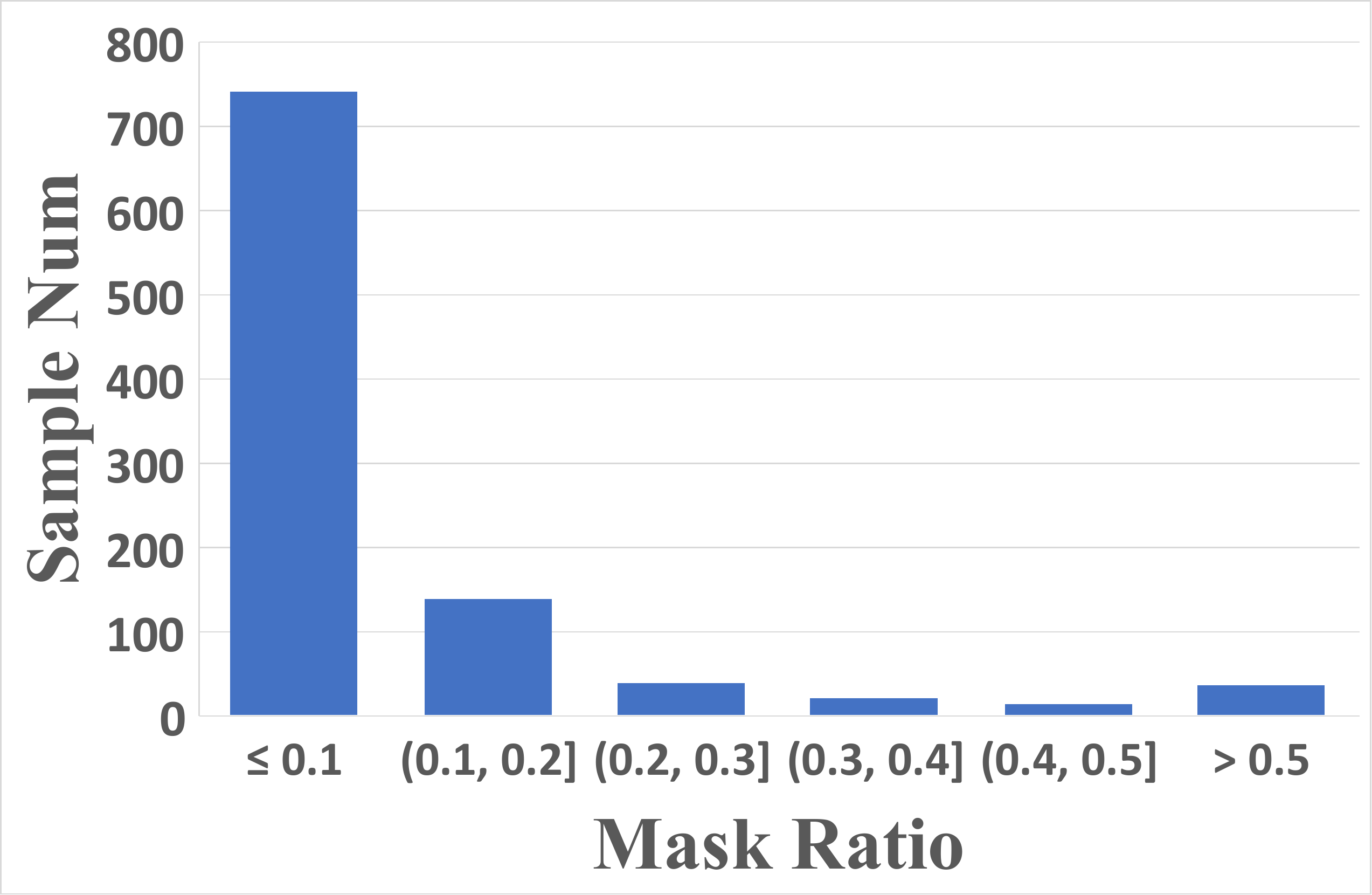}
        \caption{English to German}
        \label{fig:mask_percent:en2de} 
    \end{subfigure}
    \begin{subfigure}[t]{0.49\linewidth}
        \includegraphics[width=\linewidth]{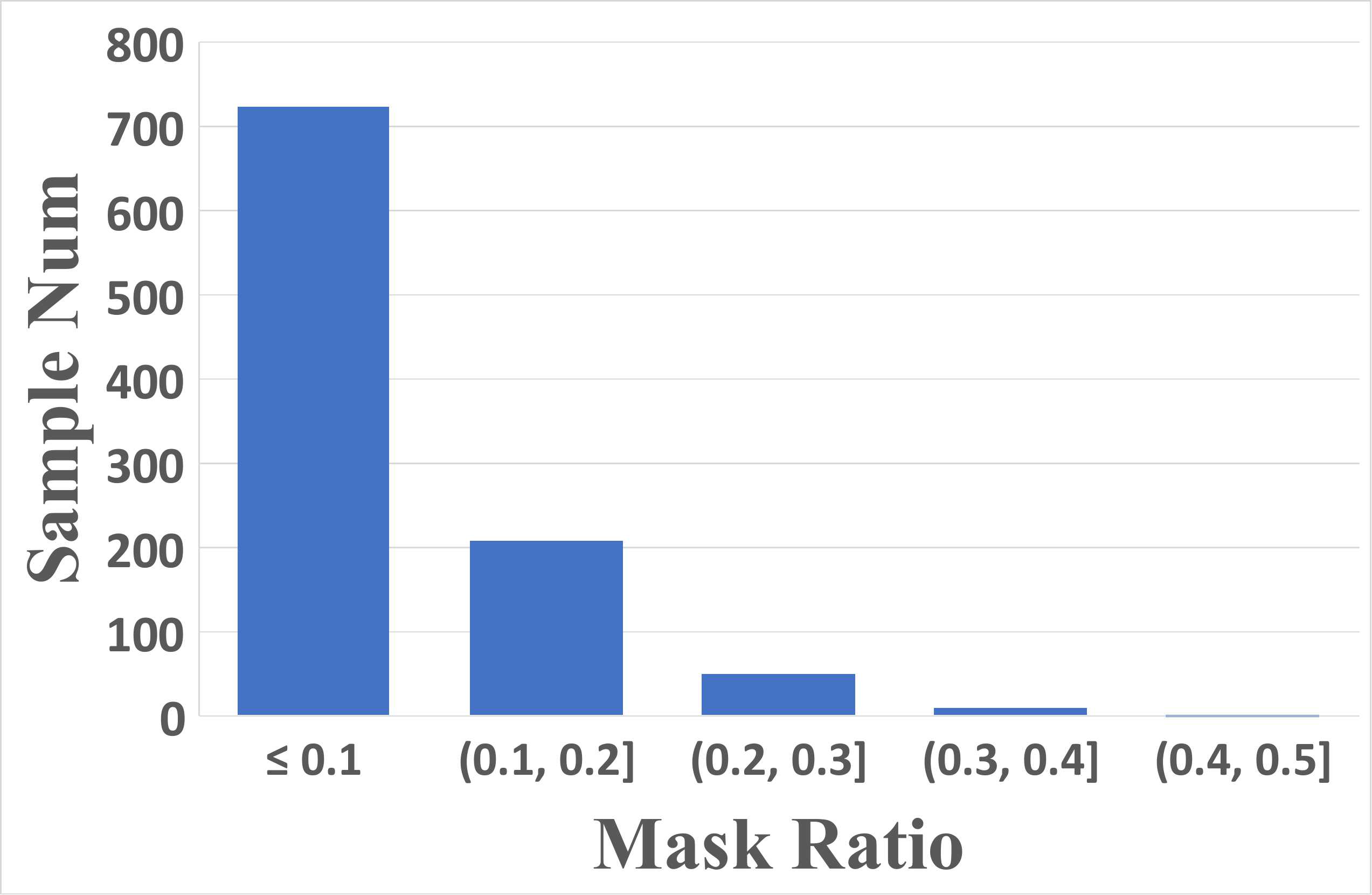}
        \caption{English to Chinese}
        \label{fig:mask_percent:en2zh} 
    \end{subfigure}
    \begin{subfigure}[t]{0.49\linewidth}
        \includegraphics[width=\linewidth]{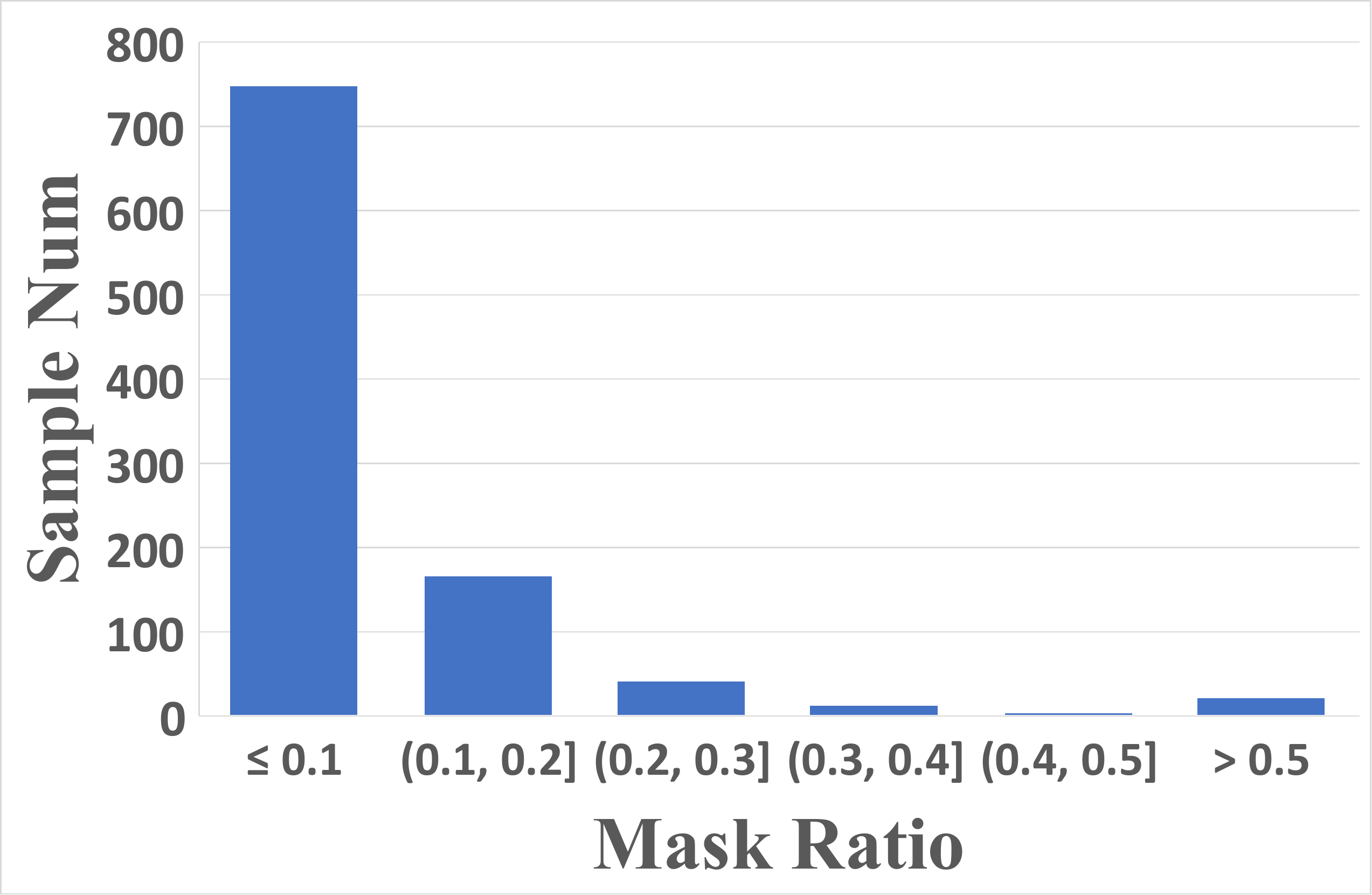}
        \caption{Chinese to English}
        \label{fig:mask_percent:zh2en} 
    \end{subfigure}
    \caption{ Histograms of the length ratios of the masked
spans on the \textit{WeTS} test sets (1000 samples each).}
    \label{fig:histgram}
\end{figure}

\begin{figure*}[ht]
    \centering
    \begin{subfigure}[t]{0.32\linewidth}
        \includegraphics[width=\linewidth]{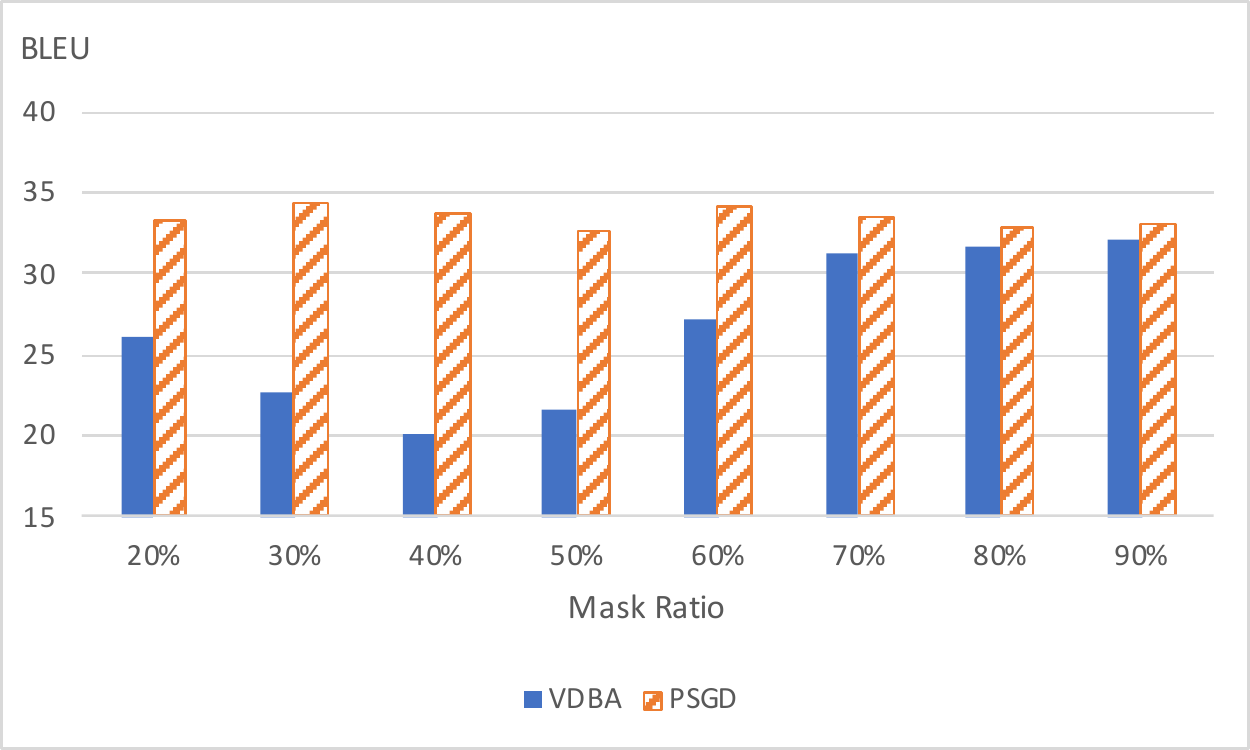}
        \caption{German to English}
        \label{fig:mask_ref_bleu:de2en} 
    \end{subfigure}
    \begin{subfigure}[t]{0.32\linewidth}
        \includegraphics[width=\linewidth]{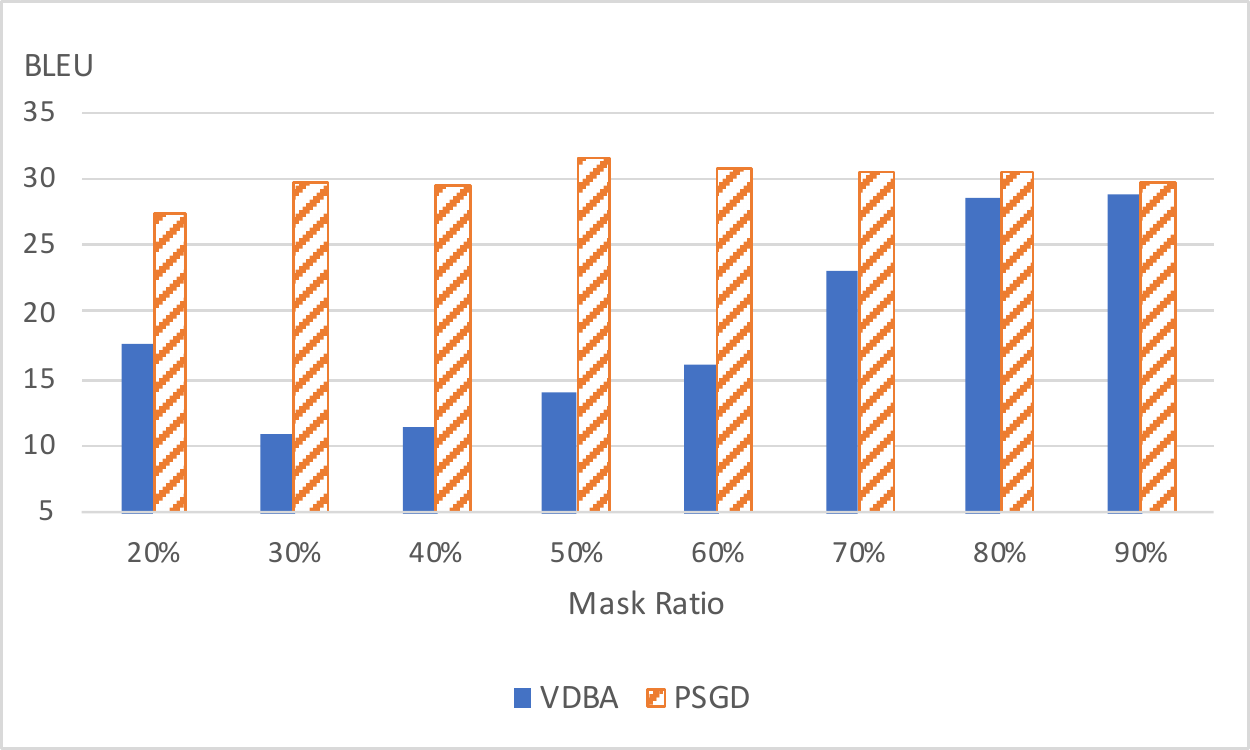}
        \caption{English to German}
        \label{fig:mask_ref_bleu:en2de} 
    \end{subfigure}
    \begin{subfigure}[t]{0.32\textwidth}
        \includegraphics[width=\linewidth]{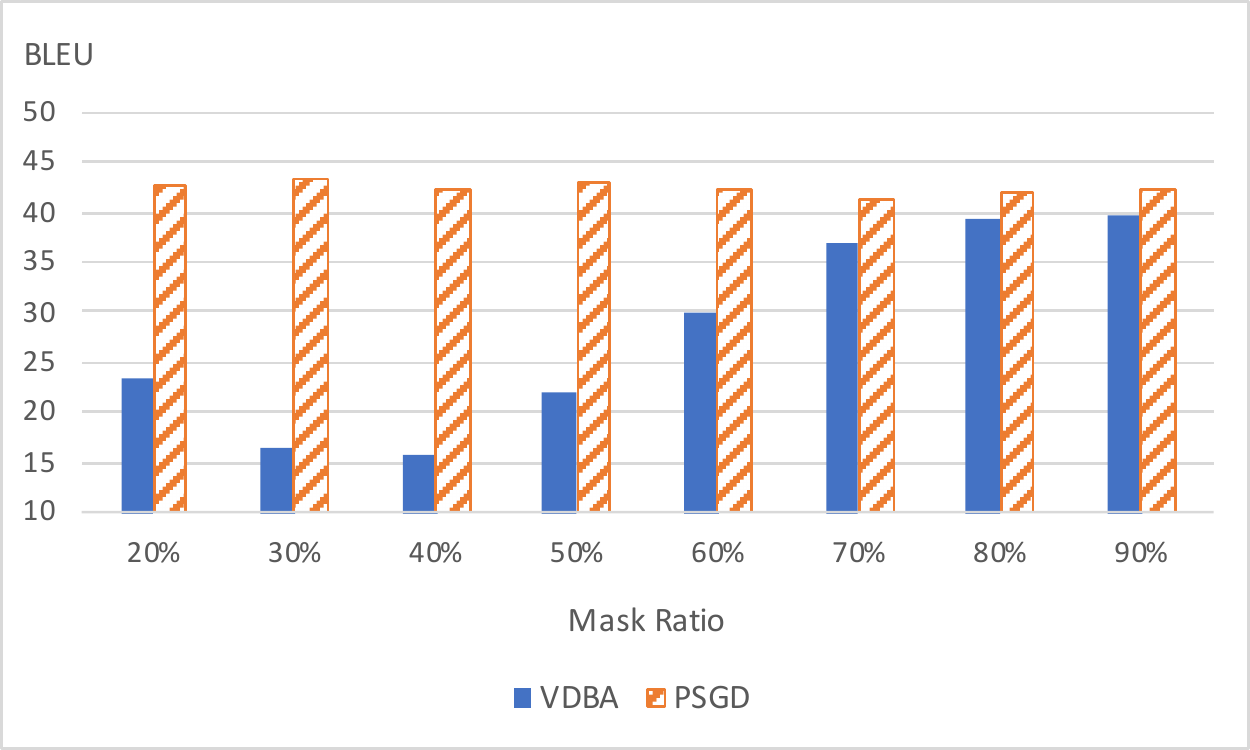}
        \caption{English to French}
        \label{fig:mask_ref_bleu:en2fr} 
    \end{subfigure}
    \caption{BLEU of the masked span with different mask ratios on WMT news test sets. PSGD outperforms VDBA in all language pairs with each mask ratios.}
    \label{fig:mask_ref_bleu}
\end{figure*}
\begin{figure*}[ht]
    \centering
    \begin{subfigure}[t]{0.32\linewidth}
        \includegraphics[width=\linewidth]{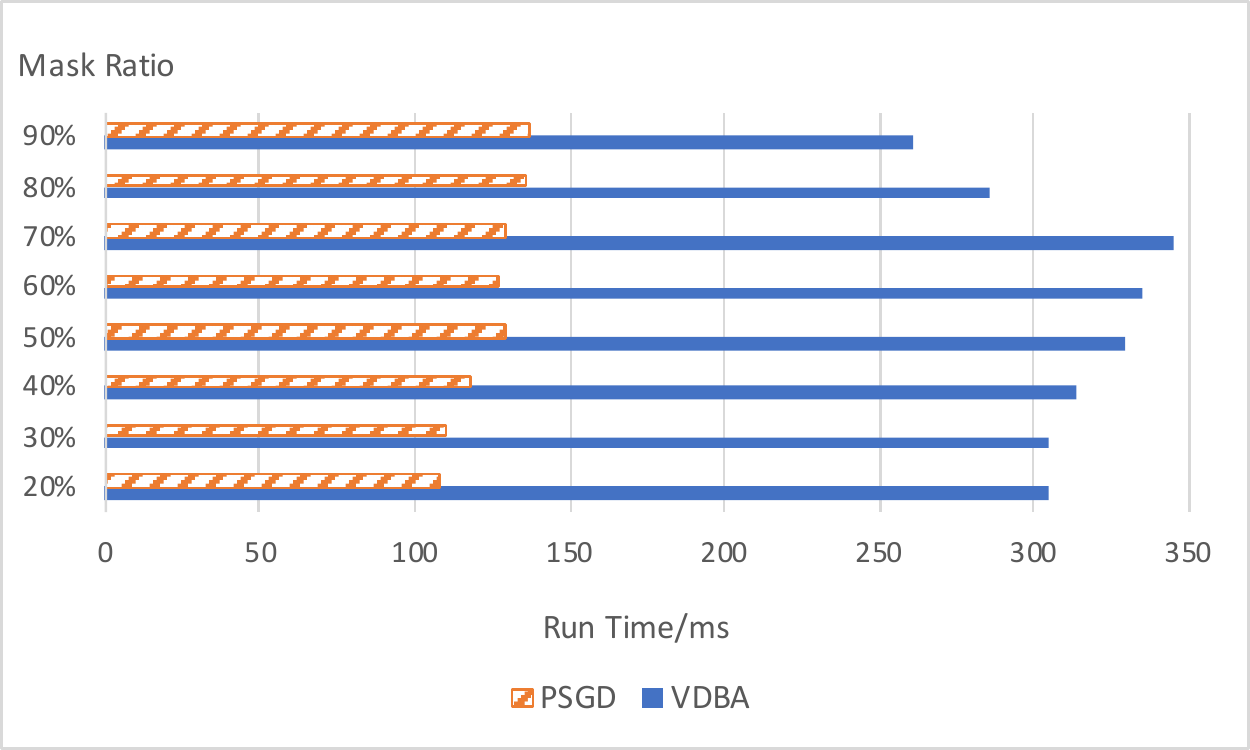}
        \caption{German to English}
        \label{fig:mask_ref_time:de2en} 
    \end{subfigure}
    \begin{subfigure}[t]{0.32\linewidth}
        \includegraphics[width=\linewidth]{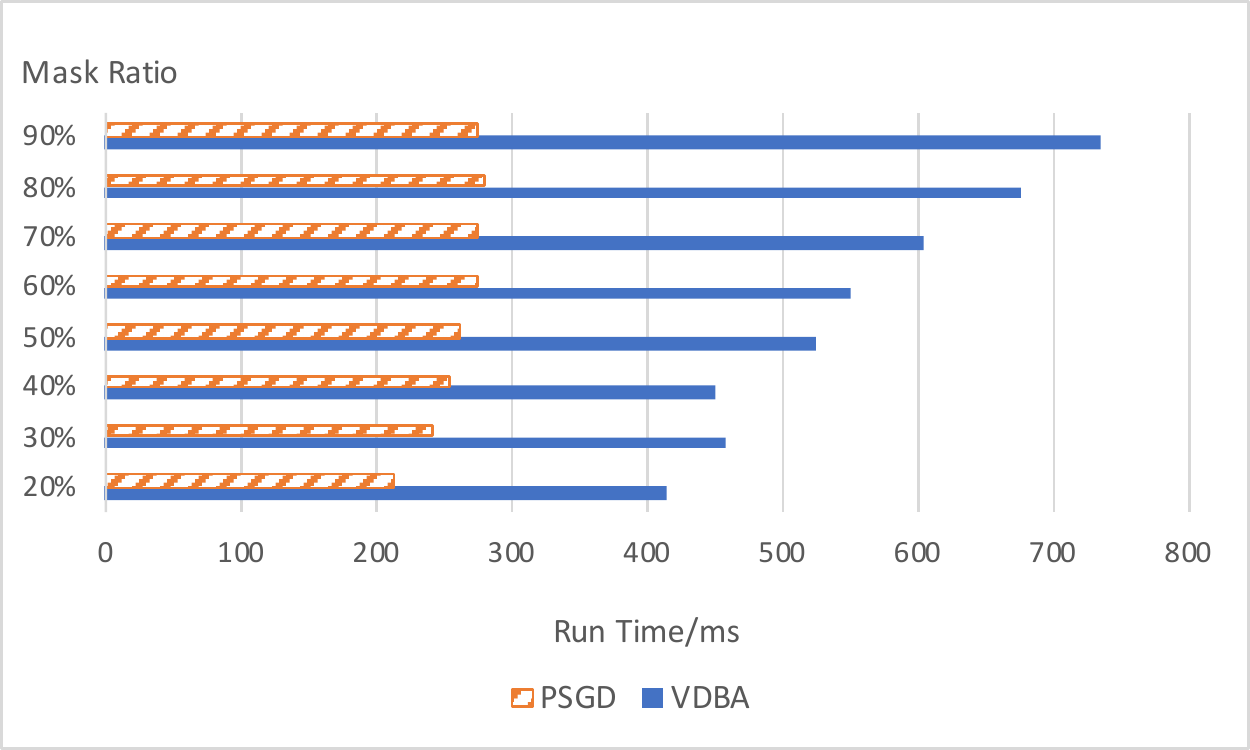}
        \caption{English to German}
        \label{fig:mask_ref_time:en2de} 
    \end{subfigure}
    \begin{subfigure}[t]{0.32\textwidth}
        \includegraphics[width=\linewidth]{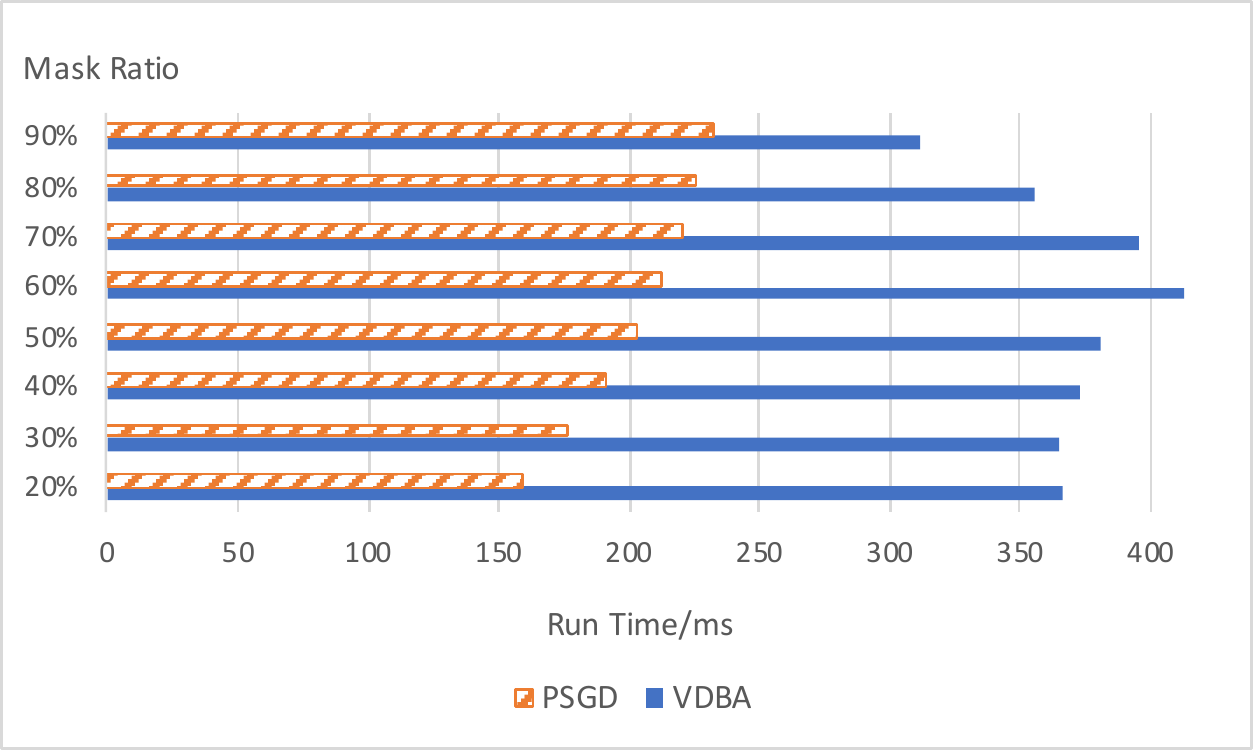}
        \caption{English to French}
        \label{fig:mask_ref_time:en2fr} 
    \end{subfigure}
    \caption{Decoding time costs with different mask ratios on WMT news test sets. PSGD saves $63.4\%$ of time costs compared with VDBA on average.}
    \label{fig:decoding_time}
\end{figure*}

\begin{figure*}[ht]
    \centering
    \begin{subfigure}[t]{0.32\linewidth}
        \includegraphics[width=\linewidth]{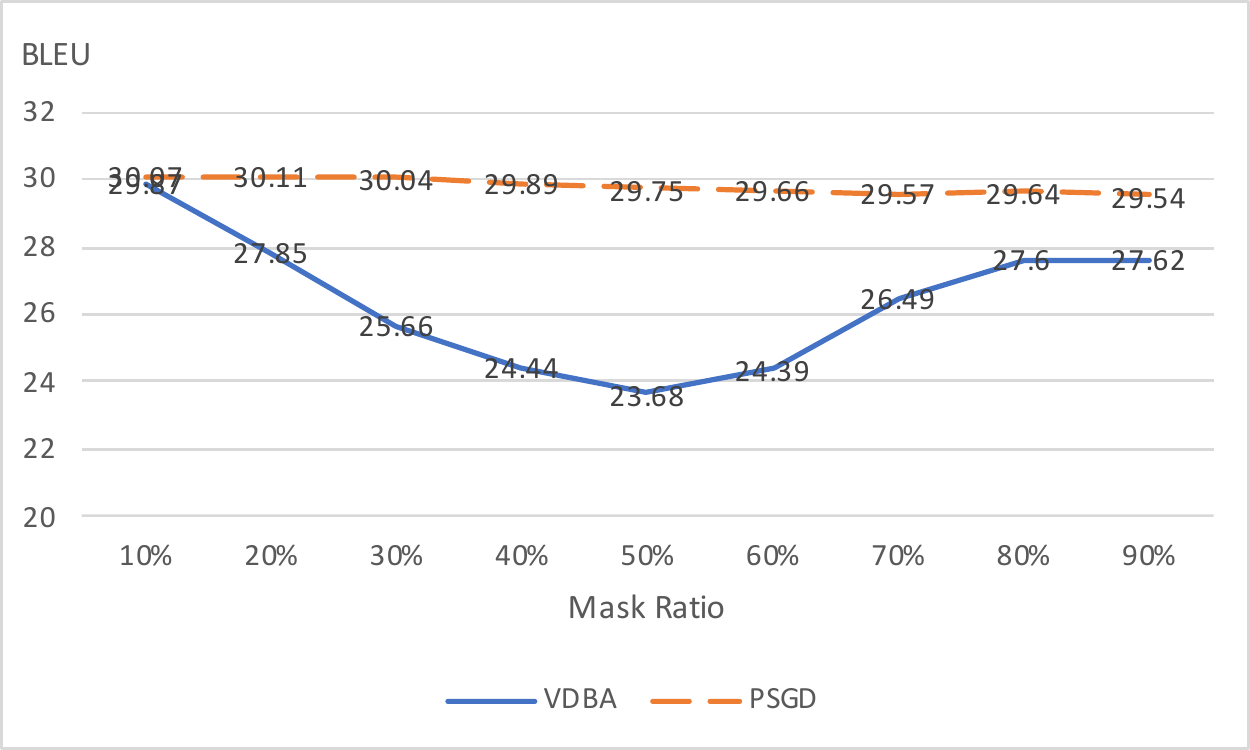}
        \caption{German to English}
        \label{fig:mask_inf_bleu:de2en} 
    \end{subfigure}
    \begin{subfigure}[t]{0.32\linewidth}
        \includegraphics[width=\linewidth]{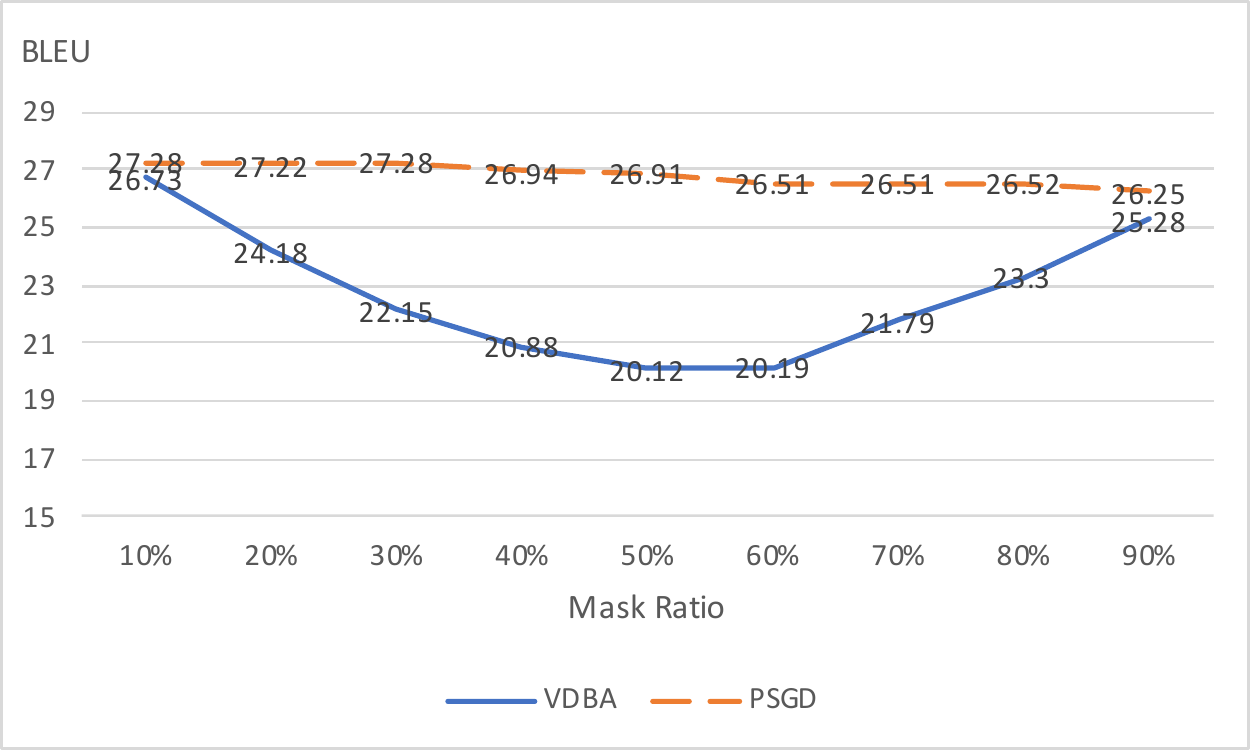}
        \caption{English to German}
        \label{fig:mask_inf_bleu:en2de} 
    \end{subfigure}
    \begin{subfigure}[t]{0.32\linewidth}
        \includegraphics[width=\linewidth]{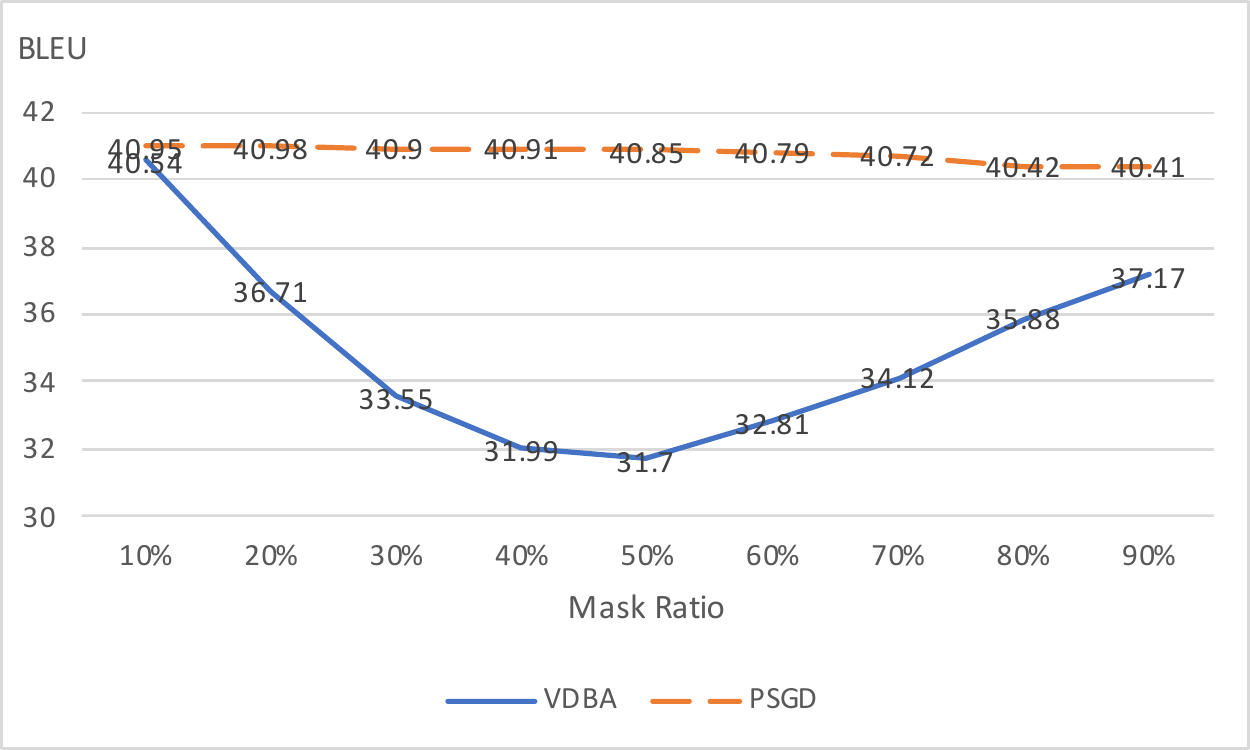}
        \caption{English to French}
        \label{fig:mask_inf_bleu:en2fr} 
    \end{subfigure}
    \caption{BLEU of whole sentences with different mask ratios when prefix and suffix constraints come from MT outputs.}
    \label{fig:MT_reconstruct}
\end{figure*}
Results on the test set of \textit{WeTS} are presented in Table \ref{tab:ts_result}. Notice that results of TSMind are not listed in Table \ref{tab:ts_result}. This is because TSMind is a system submitted to the WMT22-TS shared task. Its results on the WeTS dataset are those of systems trained with a large amount of augmented data. To demonstrate the performance of our designed unsupervised method, comparisons with models without data augmentation are more reasonable and much fairer. Therefore, we compared our model performances with the results of TSMind without data augmentation on the development set of \textit{WMT22-TS} in Table \ref{tab:wmt_result}. It can be observed that our PSGD outperforms VDBA significantly by an average of $10.87$ BLEU and an average of $8.62$ BLEU separately on the two datasets for German-English, English-German, Chinese-English, and English-Chinese. Moreover, without any extra training, PSGD is even better than the supervised SA-Transformer and TSMind systems that are trained with TS-labeled data. 
 

\subsection{Patience for Early Stopping in Decoding}
In PSGD, we involve a hyper-parameter $pt$, the number of extra decoding steps we take for early stopping. Larger $pt$ can usually avoid omissions since when a local optimum is reached, the model should be encouraged to generate in more steps to ensure that it is globally optimal. However, a larger $pt$ would slow down the decoding process. 

To determine a proper value for $pt$, we investigate the model performances in SacreBLEU and the decoding time cost on the German-English development set of $WeTS$. The results are shown in Figure \ref{fig:early_stop_fig}, which indicates that when $pt$ increases from a relatively small value, the performance of PSGD will significantly increase as well. However, there is no more much gain when $pt$ is over $5$. Taking the trade-off between time efficiency and the suggestion quality into consideration, we empirically set $pt=5$. For other language pairs, we found $pt=5$ is also relatively optimal. 

\subsection{Experiments with Different Mask Ratios}
\label{subsec:mask-bleu}
Samples in the \textit{WeTS} dataset have a short length of masked spans on average, whose masked spans contain less than $10\%$ words of the whole translation sentence as displayed in the histograms in Figure \ref{fig:histgram}. 

To fairly verify the performance of PSGD on longer masked spans, we consider applying various mask ratios on the translation datasets. We mask a span in the golden references with a random length ratio from $20\%$ to $90\%$ and obtain predictions for the spans by PSGD and VDBA separately. Then, we calculate SacreBLEU scores for comparison. 

In practice, we use the WMT 2014 English-French test set \cite{wmt14findings} and WMT 2021 English-German and German-English test sets \cite{akhbardeh-etal-2021-findings} in the news translation for evaluation.
PSGD and VDBA decode based on the publicly released NMT models, \citet{ng2019facebook} for English-German and German-English and \citet{ott2018scaling} for English-French. 

Experimental results are shown in Figure \ref{fig:mask_ref_bleu}. It shows that the proposed PSGD outperforms VDBA in all language pairs for each mask ratio. The performance of VBDA reaches the lowest point at the mask ratio of $40\%$, and then dramatically increases as the mask ratio increases. By contrast, the performance of PSGD appears more stable.

\subsection{Computation Efficiency} 
\label{subsec:mask-time}
In translation suggestion, compared with the decoding steps $t_p+t_r+t_s$ of \citet{hu-etal-2019-improved}, PSGD is more efficient with $t_r+pt$ decoding steps when $pt$ is smaller than $t_p+t_s$. 
To thoroughly evaluate the speed of PSGD's decoding process, we conduct a running-time cost analysis of translation suggestions with different mask ratios, relying on the WMT news translation test sets mentioned in Section \ref{subsec:mask-bleu}. All experiments are run on a single Tesla P-100 GPU with CUDA V10.2. The target sequences are generated with a batch size of $10$ and a beam size of $5$. We obtain the time cost by averaging the running time of each sample in the test sets. Notice that times for pre-processings, such as tokenization and subwords segmentation, are not counted because these parts are identical for both methods. Hyper-parameter $pt$ is empirically set to 5, the same value in the experiments shown in Figure \ref{fig:mask_ref_bleu}.

The time costs of PSGD and VDBA are displayed in Figure \ref{fig:decoding_time}. It is clear that PSGD runs faster than VDBA in all cases. As discussed in Section \ref{subsec:mask-bleu}, in the scene of TS, the incorrect spans are usually short, which contain less than $10\%$ of words in the whole translation sentences on average. In such a case, referred in Figure \ref{fig:decoding_time}, PSGD can speed up at least 2x times compared with VDBA.

\subsection{Robustness Test}
In the above experiments, the given prefix and suffix constraints come from the golden translation references. While in the applications of CAT, prefix and suffix constraints could include erroneous words even if the incorrect span has been selected. Therefore, it is necessary to carry out experiments where the prefix and suffix constraints contain potential errors. We assume that the performances of a robust decoding algorithm will not decrease too much in such a circumstance. 

Again, we use the machine translation datasets mentioned in Section \ref{subsec:mask-bleu} and \ref{subsec:mask-time}. The difference is that rather than masking golden references, we try to mask a span with random length in the MT result to get the prefix and suffix constraints. This setup is much closer to that of \textit{WeTS}'s. In evaluation, the Sacre-BLEU score can only be calculated between the entire golden reference and the suggestions for the masked span alone with the prefix and suffix constraints, since we don't have the parallel references for the masked spans. 

Results of PSGD and VDBA are plotted in Figure \ref{fig:MT_reconstruct}. It illustrates that PSGD significantly outperforms VDBA in all the cases of mask ratios. When the masked span is extremely short, both algorithms will naturally not impact much on the whole sequence evaluation. On the other hand, when the masked span is extremely long, the decoding process of PSGD or VDBA is almost equivalent to that of the original NMT model. Therefore, in both cases, it is unsurprising to see that the gap between the performances of PSGD and VDBA is small. 
Except for these two situations, the performance of PSGD remains at a higher level than that of VDBA. And VDBA reaches the lowest point when nearly half of MT words are masked.

\subsection{Summary of Experiments}
In this section, we conduct exhaustive experiments to evaluate the suggestion quality, time efficiency, and robustness of PSGD. PSGD significantly outperforms VDBA by an average increase of $10.87$ BLEU and $8.62$ BLEU on the benchmark datasets, \textit{WeTS} and \textit{WMT22-TS}, respectively. Experiments in time efficiency show its superiority by an overall $63.4\%$ deduction on the WMT translation datasets. In robustness tests, PSGD remains at a higher level than VDBA all the time. Finally, on the TS benchmark datasets, PSGD is superior over two supervised TS systems, SA-Transformer and TSMind, which are trained with human-annotated data.

\section{Conclusion}

In this paper, we propose a neat prefix-suffix guided decoding (PSGD) algorithm for the translation suggestion task in computer-aided translation. It emphasizes the probability of the entire generation including prefix and suffix constraints and decodes only for incorrect spans selected by human translators with an early stopping mechanism. Given a pre-trained auto-regressive NMT model, PSGD can be easily applied during inference without any additional training/fine-tuning. Comprehensive experimental results demonstrate that PSGD significantly outperforms the state-of-the-art constrained decoding algorithm, VDBA, in all of the translation quality, time efficiency, and robustness. Meanwhile, it is also superior to supervised TS systems trained with human-annotated data.

\section*{Limitations}
PSGD is a straightforward constrained-decoding algorithm for the translation suggestion task. However, the early-stopping mechanism involves extra time costs. Though PSGD is more efficient than VDBA in the scene of TS, where only two constraints (prefix and suffix) appear, it could be slower than VDBA if there were more short constraints. Besides, even if we take both prefix and suffix constraints into consideration for emphasis on the whole translation generation, the decoding process is still auto-regressive from left to right. The algorithm could be improved if we made better use of the information of suffix constraints. For example, how to apply PSGD on non-autoregressive models, such as \cite{gu-etal-2019-insertion} and \cite{yang-etal-2021-universal} will be our future work. 

\bibliography{anthology,custom}

\begin{thebibliography}{34}
\expandafter\ifx\csname natexlab\endcsname\relax\def\natexlab#1{#1}\fi

\bibitem[{Akhbardeh et~al.(2021)Akhbardeh, Arkhangorodsky, Biesialska, Bojar,
  Chatterjee, Chaudhary, Costa-jussa, Espa{\~n}a-Bonet, Fan, Federmann,
  Freitag, Graham, Grundkiewicz, Haddow, Harter, Heafield, Homan, Huck,
  Amponsah-Kaakyire, Kasai, Khashabi, Knight, Kocmi, Koehn, Lourie, Monz,
  Morishita, Nagata, Nagesh, Nakazawa, Negri, Pal, Tapo, Turchi, Vydrin, and
  Zampieri}]{akhbardeh-etal-2021-findings}
Farhad Akhbardeh, Arkady Arkhangorodsky, Magdalena Biesialska, Ond{\v{r}}ej
  Bojar, Rajen Chatterjee, Vishrav Chaudhary, Marta~R. Costa-jussa, Cristina
  Espa{\~n}a-Bonet, Angela Fan, Christian Federmann, Markus Freitag, Yvette
  Graham, Roman Grundkiewicz, Barry Haddow, Leonie Harter, Kenneth Heafield,
  Christopher Homan, Matthias Huck, Kwabena Amponsah-Kaakyire, Jungo Kasai,
  Daniel Khashabi, Kevin Knight, Tom Kocmi, Philipp Koehn, Nicholas Lourie,
  Christof Monz, Makoto Morishita, Masaaki Nagata, Ajay Nagesh, Toshiaki
  Nakazawa, Matteo Negri, Santanu Pal, Allahsera~Auguste Tapo, Marco Turchi,
  Valentin Vydrin, and Marcos Zampieri. 2021.
\newblock \href {https://aclanthology.org/2021.wmt-1.1} {Findings of the 2021
  conference on machine translation ({WMT}21)}.
\newblock In \emph{Proceedings of the Sixth Conference on Machine Translation},
  pages 1--88, Online. Association for Computational Linguistics.

\bibitem[{Bahdanau et~al.(2014)Bahdanau, Cho, and Bengio}]{bahdanau2014neural}
Dzmitry Bahdanau, Kyunghyun Cho, and Yoshua Bengio. 2014.
\newblock Neural machine translation by jointly learning to align and
  translate.
\newblock \emph{arXiv preprint arXiv:1409.0473}.

\bibitem[{Bojar et~al.(2014)Bojar, Buck, Federmann, Haddow, Koehn, Leveling,
  Monz, Pecina, Post, Saint-Amand, Soricut, Specia, and
  Tamchyna}]{wmt14findings}
Ondrej Bojar, Christian Buck, Christian Federmann, Barry Haddow, Philipp Koehn,
  Johannes Leveling, Christof Monz, Pavel Pecina, Matt Post, Herve Saint-Amand,
  Radu Soricut, Lucia Specia, and Ale\v{s} Tamchyna. 2014.
\newblock \href {http://www.aclweb.org/anthology/W/W14/W14-3302} {Findings of
  the 2014 workshop on statistical machine translation}.
\newblock In \emph{Proceedings of the Ninth Workshop on Statistical Machine
  Translation}, pages 12--58, Baltimore, Maryland, USA. Association for
  Computational Linguistics.

\bibitem[{Bowker(2002)}]{bowker2002computer}
Lynne Bowker. 2002.
\newblock \emph{Computer-aided translation technology: A practical
  introduction}.
\newblock University of Ottawa Press.

\bibitem[{Bowker(2014)}]{bowker2014computer}
Lynne Bowker. 2014.
\newblock Computer-aided translation: Translator training.
\newblock In \emph{Routledge encyclopedia of translation technology}, pages
  126--142. Routledge.

\bibitem[{Bowker and Fisher(2010)}]{bowker2010computer}
Lynne Bowker and Des Fisher. 2010.
\newblock Computer-aided translation.
\newblock \emph{Handbook of translation studies}, 1:60--65.

\bibitem[{Chatterjee(2019)}]{chatterjee2019automatic}
Rajen Chatterjee. 2019.
\newblock Automatic post-editing for machine translation.
\newblock \emph{arXiv preprint arXiv:1910.08592}.

\bibitem[{Conneau et~al.(2020)Conneau, Khandelwal, Goyal, Chaudhary, Wenzek,
  Guzm{\'a}n, Grave, Ott, Zettlemoyer, and Stoyanov}]{xlmr}
Alexis Conneau, Kartikay Khandelwal, Naman Goyal, Vishrav Chaudhary, Guillaume
  Wenzek, Francisco Guzm{\'a}n, Edouard Grave, Myle Ott, Luke Zettlemoyer, and
  Veselin Stoyanov. 2020.
\newblock \href {https://doi.org/10.18653/v1/2020.acl-main.747} {Unsupervised
  cross-lingual representation learning at scale}.
\newblock In \emph{Proceedings of the 58th Annual Meeting of the Association
  for Computational Linguistics}, pages 8440--8451, Online. Association for
  Computational Linguistics.

\bibitem[{Domingo et~al.(2016)Domingo, Peris, and
  Casacuberta}]{domingo-etal-2016-interactive}
Miguel Domingo, Alvaro Peris, and Francisco Casacuberta. 2016.
\newblock \href {https://aclanthology.org/W16-3415} {Interactive-predictive
  translation based on multiple word-segments}.
\newblock In \emph{Proceedings of the 19th Annual Conference of the {E}uropean
  Association for Machine Translation}, pages 282--291.

\bibitem[{Ge et~al.(2022)Ge, Wang, Wang, Xiao, Duan, Zhao, and
  Zhang}]{ge2022tsmind}
Xin Ge, Ke~Wang, Jiayi Wang, Nini Xiao, Xiangyu Duan, Yu~Zhao, and Yuqi Zhang.
  2022.
\newblock Tsmind: Alibaba and soochow university's submission to the wmt22
  translation suggestion task.
\newblock \emph{arXiv preprint arXiv:2211.08987}.

\bibitem[{Gonz{\'a}lez-Rubio et~al.(2016)Gonz{\'a}lez-Rubio,
  Ortiz-Mart{\'\i}nez, Casacuberta, and
  Benedi~Ruiz}]{gonzalez-rubio-etal-2016-beyond}
Jes{\'u}s Gonz{\'a}lez-Rubio, Daniel Ortiz-Mart{\'\i}nez, Francisco
  Casacuberta, and Jos{\'e}~Miguel Benedi~Ruiz. 2016.
\newblock \href {https://doi.org/10.18653/v1/K16-1020} {Beyond prefix-based
  interactive translation prediction}.
\newblock In \emph{Proceedings of the 20th {SIGNLL} Conference on Computational
  Natural Language Learning}, pages 198--207, Berlin, Germany. Association for
  Computational Linguistics.

\bibitem[{Green et~al.(2014)Green, Chuang, Heer, and
  Manning}]{green2014predictive}
Spence Green, Jason Chuang, Jeffrey Heer, and Christopher~D Manning. 2014.
\newblock Predictive translation memory: A mixed-initiative system for human
  language translation.
\newblock In \emph{Proceedings of the 27th annual ACM symposium on User
  interface software and technology}, pages 177--187.

\bibitem[{Green et~al.(2015)Green, Heer, and Manning}]{green2015natural}
Spence Green, Jeffrey Heer, and Christopher~D Manning. 2015.
\newblock Natural language translation at the intersection of ai and hci.
\newblock \emph{Communications of the ACM}, 58(9):46--53.

\bibitem[{Gu et~al.(2019)Gu, Liu, and Cho}]{gu-etal-2019-insertion}
Jiatao Gu, Qi~Liu, and Kyunghyun Cho. 2019.
\newblock \href {https://doi.org/10.1162/tacl_a_00292} {Insertion-based
  decoding with automatically inferred generation order}.
\newblock \emph{Transactions of the Association for Computational Linguistics},
  7:661--676.

\bibitem[{Herbig et~al.(2020)Herbig, D{\"u}wel, Pal, Meladaki, Monshizadeh,
  Kr{\"u}ger, and van Genabith}]{herbig2020mmpe}
Nico Herbig, Tim D{\"u}wel, Santanu Pal, Kalliopi Meladaki, Mahsa Monshizadeh,
  Antonio Kr{\"u}ger, and Josef van Genabith. 2020.
\newblock Mmpe: A multi-modal interface for post-editing machine translation.
\newblock In \emph{Proceedings of the 58th Annual Meeting of the Association
  for Computational Linguistics}, pages 1691--1702.

\bibitem[{Hu et~al.(2019)Hu, Khayrallah, Culkin, Xia, Chen, Post, and
  Van~Durme}]{hu-etal-2019-improved}
J.~Edward Hu, Huda Khayrallah, Ryan Culkin, Patrick Xia, Tongfei Chen, Matt
  Post, and Benjamin Van~Durme. 2019.
\newblock \href {https://doi.org/10.18653/v1/N19-1090} {Improved lexically
  constrained decoding for translation and monolingual rewriting}.
\newblock In \emph{Proceedings of the 2019 Conference of the North {A}merican
  Chapter of the Association for Computational Linguistics: Human Language
  Technologies, Volume 1 (Long and Short Papers)}, pages 839--850, Minneapolis,
  Minnesota. Association for Computational Linguistics.

\bibitem[{Knowles and Koehn(2016)}]{knowles-koehn-2016-neural}
Rebecca Knowles and Philipp Koehn. 2016.
\newblock \href {https://aclanthology.org/2016.amta-researchers.9} {Neural
  interactive translation prediction}.
\newblock In \emph{Conferences of the Association for Machine Translation in
  the Americas: MT Researchers' Track}, pages 107--120, Austin, TX, USA. The
  Association for Machine Translation in the Americas.

\bibitem[{Koehn(2009)}]{koehn2009statistical}
Philipp Koehn. 2009.
\newblock \emph{Statistical machine translation}.
\newblock Cambridge University Press.

\bibitem[{L{\"a}ubli et~al.(2013)L{\"a}ubli, Fishel, Massey, Ehrensberger-Dow,
  Volk, O'Brien, Simard, and Specia}]{laubli2013assessing}
Samuel L{\"a}ubli, Mark Fishel, Gary Massey, Maureen Ehrensberger-Dow, Martin
  Volk, Sharon O'Brien, Michel Simard, and Lucia Specia. 2013.
\newblock Assessing post-editing efficiency in a realistic translation
  environment.

\bibitem[{Lengyel et~al.(2004)Lengyel, István, Kis, Balázs, Ugray, and
  Gábor}]{lengyelmemoq}
Lengyel, István, Kis, Balázs, Ugray, and Gábor. 2004.
\newblock Memoq: A new approach to computer-assisted translation.

\bibitem[{Lopez(2008)}]{lopez2008statistical}
Adam Lopez. 2008.
\newblock Statistical machine translation.
\newblock \emph{ACM Computing Surveys (CSUR)}, 40(3):1--49.

\bibitem[{Ng et~al.(2019)Ng, Yee, Baevski, Ott, Auli, and
  Edunov}]{ng2019facebook}
Nathan Ng, Kyra Yee, Alexei Baevski, Myle Ott, Michael Auli, and Sergey Edunov.
  2019.
\newblock Facebook fair's wmt19 news translation task submission.
\newblock \emph{arXiv preprint arXiv:1907.06616}.

\bibitem[{Ott et~al.(2019)Ott, Edunov, Baevski, Fan, Gross, Ng, Grangier, and
  Auli}]{ott2019fairseq}
Myle Ott, Sergey Edunov, Alexei Baevski, Angela Fan, Sam Gross, Nathan Ng,
  David Grangier, and Michael Auli. 2019.
\newblock fairseq: A fast, extensible toolkit for sequence modeling.
\newblock In \emph{Proceedings of NAACL-HLT 2019: Demonstrations}.

\bibitem[{Ott et~al.(2018)Ott, Edunov, Grangier, and Auli}]{ott2018scaling}
Myle Ott, Sergey Edunov, David Grangier, and Michael Auli. 2018.
\newblock \href {http://arxiv.org/abs/1806.00187} {Scaling neural machine
  translation}.
\newblock \emph{CoRR}, abs/1806.00187.

\bibitem[{Pal et~al.(2016)Pal, Zampieri, Naskar, Nayak, Vela, and van
  Genabith}]{pal2016catalog}
Santanu Pal, Marcos Zampieri, Sudip~Kumar Naskar, Tapas Nayak, Mihaela Vela,
  and Josef van Genabith. 2016.
\newblock Catalog online: Porting a post-editing tool to the web.
\newblock In \emph{Proceedings of the Tenth International Conference on
  Language Resources and Evaluation (LREC'16)}, pages 599--604.

\bibitem[{Peris et~al.(2017)Peris, Domingo, and
  Casacuberta}]{peris2017interactive}
{\'A}lvaro Peris, Miguel Domingo, and Francisco Casacuberta. 2017.
\newblock Interactive neural machine translation.
\newblock \emph{Computer Speech \& Language}, 45:201--220.

\bibitem[{Post(2018)}]{post2018call}
Matt Post. 2018.
\newblock A call for clarity in reporting bleu scores.
\newblock \emph{arXiv preprint arXiv:1804.08771}.

\bibitem[{Post and Vilar(2018)}]{post-vilar-2018-fast}
Matt Post and David Vilar. 2018.
\newblock \href {https://doi.org/10.18653/v1/N18-1119} {Fast lexically
  constrained decoding with dynamic beam allocation for neural machine
  translation}.
\newblock In \emph{Proceedings of the 2018 Conference of the North {A}merican
  Chapter of the Association for Computational Linguistics: Human Language
  Technologies, Volume 1 (Long Papers)}, pages 1314--1324, New Orleans,
  Louisiana. Association for Computational Linguistics.

\bibitem[{Santy et~al.(2019)Santy, Dandapat, Choudhury, and
  Bali}]{santy2019inmt}
Sebastin Santy, Sandipan Dandapat, Monojit Choudhury, and Kalika Bali. 2019.
\newblock Inmt: Interactive neural machine translation prediction.
\newblock In \emph{Proceedings of the 2019 Conference on Empirical Methods in
  Natural Language Processing and the 9th International Joint Conference on
  Natural Language Processing (EMNLP-IJCNLP): System Demonstrations}, pages
  103--108.

\bibitem[{Sutskever et~al.(2014)Sutskever, Vinyals, and
  Le}]{sutskever2014sequence}
Ilya Sutskever, Oriol Vinyals, and Quoc~V Le. 2014.
\newblock Sequence to sequence learning with neural networks.
\newblock In \emph{Advances in neural information processing systems}, pages
  3104--3112.

\bibitem[{Vaswani et~al.(2017)Vaswani, Shazeer, Parmar, Uszkoreit, Jones,
  Gomez, Kaiser, and Polosukhin}]{vaswani2017attention}
Ashish Vaswani, Noam Shazeer, Niki Parmar, Jakob Uszkoreit, Llion Jones,
  Aidan~N Gomez, {\L}ukasz Kaiser, and Illia Polosukhin. 2017.
\newblock Attention is all you need.
\newblock In \emph{Advances in Neural Information Processing Systems}, pages
  5998--6008.

\bibitem[{Yang et~al.(2022)Yang, Meng, Zhang, Li, and
  Zhou}]{yang-EtAl:2022:WMT1}
Zhen Yang, Fandong Meng, Yingxue Zhang, Ernan Li, and Jie Zhou. 2022.
\newblock \href {https://aclanthology.org/2022.wmt-1.76} {Findings of the wmt
  2022 shared task on translation suggestion}.
\newblock In \emph{Proceedings of the Seventh Conference on Machine
  Translation}, pages 821--829, Abu Dhabi. Association for Computational
  Linguistics.

\bibitem[{Yang et~al.(2021{\natexlab{a}})Yang, Zhang, Li, Meng, and
  Zhou}]{yang2021wets}
Zhen Yang, Yingxue Zhang, Ernan Li, Fandong Meng, and Jie Zhou.
  2021{\natexlab{a}}.
\newblock Wets: A benchmark for translation suggestion.
\newblock \emph{arXiv preprint arXiv:2110.05151}.

\bibitem[{Yang et~al.(2021{\natexlab{b}})Yang, Yang, Cer, Law, and
  Darve}]{yang-etal-2021-universal}
Ziyi Yang, Yinfei Yang, Daniel Cer, Jax Law, and Eric Darve.
  2021{\natexlab{b}}.
\newblock \href {https://doi.org/10.18653/v1/2021.emnlp-main.502} {Universal
  sentence representation learning with conditional masked language model}.
\newblock In \emph{Proceedings of the 2021 Conference on Empirical Methods in
  Natural Language Processing}, pages 6216--6228, Online and Punta Cana,
  Dominican Republic. Association for Computational Linguistics.

\end{thebibliography}
\bibliographystyle{acl_natbib}




\end{document}